\documentclass{article}

\usepackage[nonatbib]{neurips_data_2023}

\usepackage[utf8]{inputenc} 
\usepackage[T1]{fontenc}    
\usepackage{hyperref}       
\usepackage{url}            
\usepackage{booktabs}       
\usepackage{csquotes}
\usepackage{amsmath}

\usepackage{amsfonts}       
\usepackage{nicefrac}       
\usepackage{microtype}      
\usepackage{xcolor}         
\usepackage{graphicx} 
\usepackage{booktabs} 
\usepackage{xparse}   
\usepackage[normalem]{ulem} 
\usepackage{multirow, tabularx}
\usepackage{subcaption}

\title{EHRSHOT: An EHR Benchmark for Few-Shot Evaluation of Foundation Models}

\author{%
  Michael Wornow\thanks{Equal contribution} \\
  Department of Computer Science\\
  Stanford University\\
  \texttt{mwornow@stanford.edu} \\
  \And
  Rahul Thapa\footnotemark[1] \\
  Center for Biomedical Informatics Research\\
  Stanford University\\
  \texttt{rthapa84@stanford.edu} \\
  \And
  Ethan Steinberg \\
  Department of Computer Science\\
  Stanford University\\
  \texttt{ethan@stanford.edu} \\
  \And
  Jason A. Fries\thanks{Equal senior authorship} \\
  Center for Biomedical Informatics Research\\
  Stanford University\\
  \texttt{jfries@stanford.edu} \\
  \And
  Nigam H. Shah\footnotemark[2] \\
 Center for Biomedical Informatics Research\\
 Clinical Excellence Research Center\\
  Stanford University\\
 Technology and Digital Solutions\\
 Stanford Healthcare\\
  \texttt{nigam@stanford.edu} \\
}

\begin{document}

\maketitle

\begin{abstract}
    While the general machine learning (ML) community has benefited from public datasets, tasks, and models, the progress of ML in healthcare has been hampered by a lack of such shared assets. The success of foundation models creates new challenges for healthcare ML by requiring access to shared pretrained models to validate performance benefits. We help address these challenges through three contributions. First, we publish a new dataset, {\sc EHRSHOT}, which contains de-identified structured data from the electronic health records (EHRs) of 6,739 patients from Stanford Medicine. Unlike MIMIC-III/IV and other popular EHR datasets, {\sc EHRSHOT} is longitudinal and not restricted to ICU/ED patients. Second, we publish the weights of CLMBR-T-base, a 141M parameter clinical foundation model pretrained on the structured EHR data of 2.57M patients. We are one of the first to fully release such a model for coded EHR data; in contrast, most prior models released for clinical data  (e.g. GatorTron, ClinicalBERT) only work with unstructured text and cannot process the rich, structured data within an EHR. We provide an end-to-end pipeline for the community to validate and build upon its performance. Third, we define 15 few-shot clinical prediction tasks, enabling evaluation of foundation models on benefits such as sample efficiency and task adaptation. Our model and dataset are available via a research data use agreement from \href{https://ehrshot.stanford.edu/}{our website}. Code to reproduce our results is available \href{https://github.com/som-shahlab/ehrshot-benchmark}{here}.
\end{abstract}

%
%

\section{Introduction}
\label{section_intro}

Open datasets, code, and models have been essential in advancing machine learning (ML) over the past decade \cite{russakovsky2015imagenet, toscher2009bigchaos,langenkamp2022open}. Though the benefits of open code and data are well known \cite{sonnenburg2007need, piwowar2013data}, there is currently a dearth of publicly available datasets and pretrained models for electronic health records (EHRs), which makes conducting reproducible research challenging \cite{sohn2023reproducibility,mcdermott2021reproducibility}.

This is especially problematic in the era of foundation models (FMs), which hold tremendous promise for clinical applications \cite{moor2023foundation}. The ability of a shared FM to generalize across health systems would be highly valuable, as most hospitals lack the computational resources to train such models \cite{sevilla2022compute}. Yet many of the purported benefits of clinical FMs, such as sample efficiency and task adaptability, remain difficult to evaluate due to reproducibility and data access issues \cite{sohn2023reproducibility}.

Unfortunately, most existing EHR datasets (e.g., MIMIC-III/IV \cite{johnson2016mimic, johnson2023mimic}, eICU \cite{pollard2018eicu}, AmsterdamUMCdb \cite{thoral2021sharing}, and HiRID \cite{faltys10hirid}) narrowly focus on the intensive care unit (ICU), which provides a limited snapshot of a patient's overall health trajectory and limits what tasks can be evaluated \cite{van2023yet}. Access to a patient's complete medical timeline, referred to as "longitudinal" data, offers a more realistic representation of the breadth of information available to a health system. Longitudinal EHR data, however, remains scarce. The few public datasets that exist, such as the CPRD \cite{herrett2015data} and UK BioBank \cite{bycroft2018uk}, lack consensus on shared evaluation tasks / data processing pipelines and require navigating a research protocol review process, which creates challenges when curating shared ML workflows \cite{tang2020democratizing}.

While the limitations of prior benchmarks were less apparent when developing small-scale, task-specific models, their utility is limited for evaluating FMs on task adaptation, few-shot learning, and other properties of large-scale, self-supervised models \cite{opportunities_and_risks, qiu2023large}. Clinical FMs surface new questions, and a dataset for evaluating such FMs should contain a diverse range of tasks in low-label settings with longitudinal data \cite{mcdermott2021comprehensive}. Most importantly, such a benchmark should also release the weights of its pretrained models so the community can reproduce and build upon its results. Unfortunately, few FMs trained on EHR data have had their model weights published \cite{wornow2023shaky}.

Our work helps address both shortcomings -- a lack of public EHR datasets and pretrained clinical FMs -- as one of the first combined releases of a research dataset and FM trained on EHR data. We outline our three primary contributions towards more reproducible ML for healthcare below:

\begin{enumerate}
    \item We release {\sc EHRSHOT}, a longitudinal EHR benchmark for the few-shot evaluation of clinical FMs. {\sc EHRSHOT} contains the \textbf{full coded medical timelines of 6,739 patients} from Stanford Medicine. Records include demographics, diagnoses, procedures, laboratory results, medications, and other structured data, for a total of 41.6 million clinical events across 921,499 encounters. {\sc EHRSHOT} contains an average of \textbf{2.3x more clinical events and 95.2x more encounters per patient than MIMIC-IV \cite{johnson2023mimic}} and, unlike the majority of existing benchmarks, includes patients not seen in the ICU or emergency department (ED).
    \item We publish the weights of a \textbf{141M parameter transformer-based foundation model} (CLMBR-T-base) pretrained on the deidentified structured data of \textbf{2.57M patients' EHRs}. CLMBR-T-base was trained in a self-supervised manner to autoregressively predict the next code in a patient's timeline given their previous codes \cite{steinberg2021language}. We are among the first to publish the full weights of such a clinical FM \cite{wornow2023shaky} for the community to evaluate and build upon. Researchers who leverage our model can benefit from both improved downstream task accuracy and cost savings by shortcutting the model development process.
    \item We define a new few-shot benchmark of \textbf{15 patient classification tasks.} Several tasks have naturally low prevalence, creating a realistic setting for few-shot experimentation. While our pretrained model offers significant AUROC/AUPRC gains in few-shot settings over a traditional supervised baseline, we demonstrate that there remains significant room for improvement on many of our tasks. 
\end{enumerate}

\begin{figure}[t]
    \centering
    \includegraphics[scale=0.3]{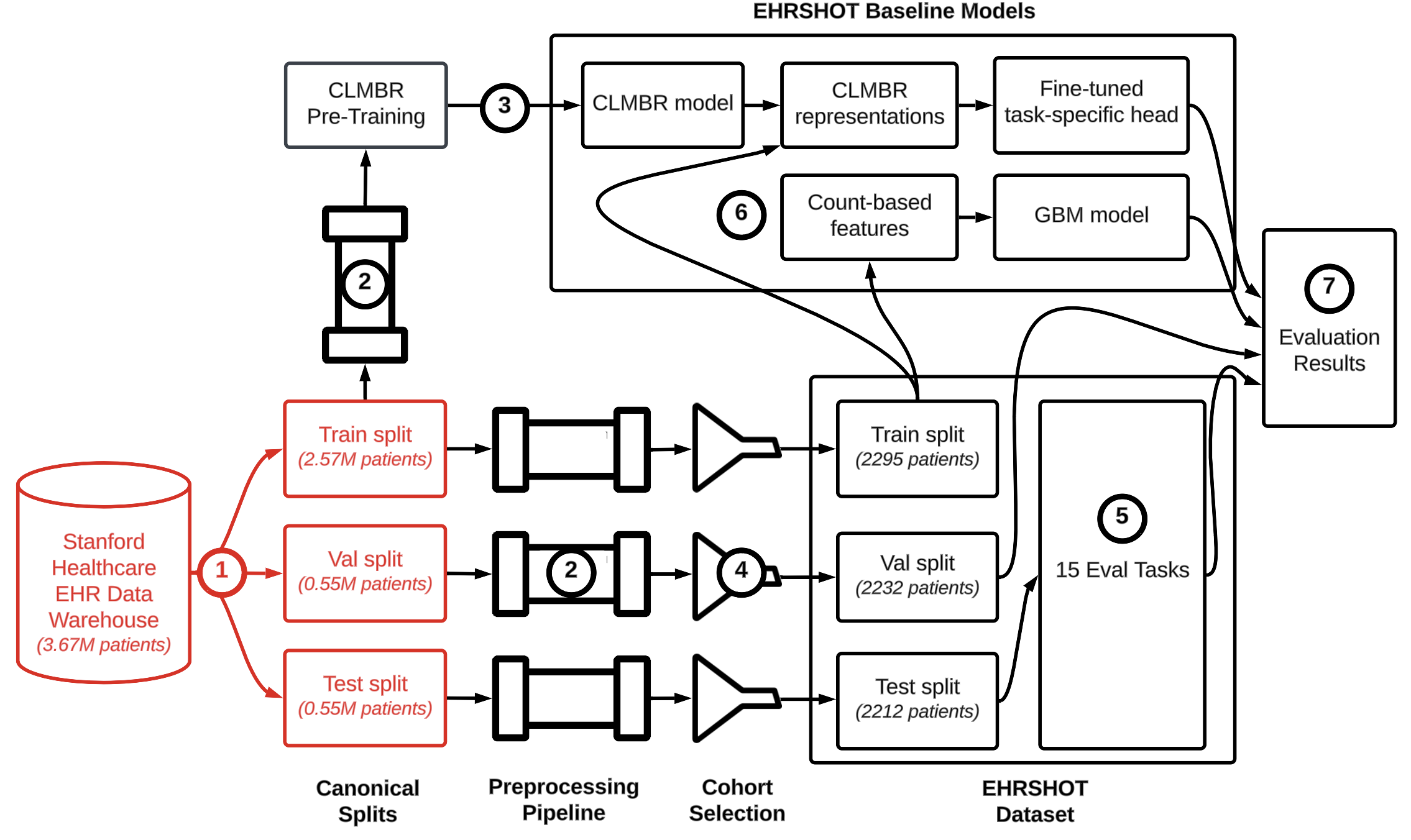}
    \caption{Overview of {\sc EHRSHOT}. \fbox{\textcolor{black}{Black boxes}} represent open source code, data, and model weights. \fbox{\textcolor{red}{Red boxes}} are private data. (1) Starting with a source EHR database of 3.67M patients, we define a global train/val/test split across all patients. (2) We use an \href{https://github.com/som-shahlab/femr}{open source EHR preprocessing package called FEMR} to transform our data. We keep all structured data (diagnoses, medications, labs, etc.) and discard images and clinical text. (3) We use the 2.57M patients in our global train split to pre-train a foundation model, CLMBR-T-base \cite{steinberg2021language} (4) We filter the source database down to a cohort of 6,739 patients, which we use for {\sc EHRSHOT}. (5) We define 15 few-shot classification tasks and label each patient accordingly. (6) We test two baseline models for each task: our pretrained CLMBR-T-base and a count-based GBM model \cite{rajkomar2018scalable}. (7) We measure the AUROC and AUPRC of each model on each task, and share the results in Section \ref{section_results}.}
    \label{fig:figure_1}
\end{figure}

Our overall workflow is shown in Figure \ref{fig:figure_1}. We publish the full code to replicate our results here: \href{https://github.com/som-shahlab/ehrshot-benchmark}{https://github.com/som-shahlab/ehrshot-benchmark}. We also publish the full weights of our pretrained clinical foundation model, as well as the {\sc EHRSHOT} dataset and task labels, under a non-commercial data usage agreement here: \href{https://ehrshot.stanford.edu}{https://ehrshot.stanford.edu}.

%
%

\section{Related Work}
\label{section_related_work}

One of the most popular EHR datasets made accessible to researchers is MIMIC-III, which contains roughly 40,000 patients seen in the intensive care unit (ICU) of Beth Israel Deaconess Medical Center in Boston, Massachusetts, between 2001 and 2012 \cite{johnson2016mimic}. Other public datasets include eICU \cite{pollard2018eicu}, HiRID \cite{faltys10hirid}, AmsterdamUMCdb \cite{thoral2021sharing}, CPRD \cite{herrett2015data}, MIMIC-IV \cite{johnson2023mimic}, and the UK BioBank \cite{bycroft2018uk}.

Most of the aforementioned datasets are narrowly scoped to a single department: the ICU \cite{johnson2016mimic, pollard2018eicu, faltys10hirid, thoral2021sharing}. This makes it impossible to capture a patient's full health trajectory to the extent that an academic medical center or health system would know of the patients it treats. Other datasets such as MIMIC-IV include data from multiple departments, but are still heavily anchored to the ICU, as only patients admitted for an ICU/ED visit are included \cite{johnson2023mimic}. In contrast, our work releases the full longitudinal EHR of patients across all departments of a major academic medical center, thus providing a more realistic setting for general prediction making.

Prior work has also typically relied on the creation of bespoke schemas to store their data. These custom schemas greatly increase the difficulty of transferring models across datasets and sites \cite{tang2020democratizing}. In contrast, the data preprocessing pipeline that we use is capable of ingesting both {\sc EHRSHOT} as well as any dataset that follows the Observational Medical Outcomes Partnership Common Data Model (OMOP-CDM), an open community data standard for sharing EHRs used by over 100 health systems \cite{omop-cdm}. More details on our data preprocessing pipeline can be found in the Appendix in Section \ref{section_data_preprocessing}.

Previously published EHR datasets typically only provide raw data. Thus, significant additional effort has been devoted to building standardized preprocessing pipelines, patient splits, and task definitions on top of these datasets \cite{harutyunyan2019multitask, purushotham2018benchmarking, mcdermott2021reproducibility}. These add-on benchmarks, however, are still limited by the narrow scope of their underlying data, and many recycle the same core set of tasks (e.g. in-patient mortality, long length-of-stay, ICU transfer, and ICD code prediction) \cite{purushotham2018benchmarking, harutyunyan2019multitask, gupta2022extensive}. Additionally, these benchmarks are typically not created with the purpose of measuring a pretrained model's few-shot performance \cite{mcdermott2021comprehensive}. This limits their utility in assessing the key value propositions of foundation models, such as improved sample efficiency and adaptation to diverse tasks. 

On the modeling side, substantial literature exists on training FMs for EHR data \cite{poulain2022few, li2020behrt, rasmy2021med, steinberg2021language, munoz2022sehr}. However, the vast majority of these FMs have never had their weights published \cite{wornow2023shaky}. This greatly hinders reproducibility and makes cross-model evaluations difficult. Worse, this lack of sharing undermines a primary advantage of FMs: transfer learning, i.e. the ability to use the pretrained weights of an existing FM to shortcut model development for other tasks \cite{opportunities_and_risks}. 

{\sc EHRSHOT} aims to fill several of these gaps by providing a longitudinal EHR benchmark specifically geared towards few-shot evaluation of pretrained FMs. {\sc EHRSHOT} is built on top of a cross-site interoperable standard (OMOP-CDM), and leverages an open source data preprocessing pipeline to allow other researchers to reproduce our results end-to-end. Additionally, we release the weights of the clinical foundation model that we pretrain and evaluate, one of the first to do so. We provide additional points of comparison in Table \ref{tab:prior_work}.

\newcolumntype{P}[1]{>{\centering\arraybackslash}p{#1}}
\begin{table}[h]

    \scriptsize
    \centering
    \caption{Comparison of our work to existing EHR benchmarks. Checkmark indicates full support, asterisk represents properties that are semi-supported.}
    
    \vspace{0.2cm}
    
    \begin{tabular}{l | p{2cm} | P{0.65cm}P{0.5cm}P{0.65cm} | P{0.4cm}P{0.4cm} | P{0.9cm}P{1.3cm}P{0.75cm}}

        \toprule
        
        \multirow{3}{*}{\textbf{Benchmark}} & \multicolumn{1}{c}{\textbf{Source} } & \multicolumn{3}{c}{\textbf{EHR Properties }} & \multicolumn{2}{c}{\textbf{Evaluation}} & \multicolumn{3}{c}{\textbf{Reproducibility }}\\

         & \multirow{2}{=}{\centering Dataset} & \multirow{2}{=}{\centering ICU/ED Visits} & \multirow{2}{=}{\centering Other Visits} & \multirow{2}{=}{\centering \# of Patients} &\multirow{2}{=}{\centering \# of Tasks} & \multirow{2}{=}{\centering Few Shot} & \multirow{2}{=}{\centering Dataset via DUA} & \multirow{2}{=}{\centering Preprocessing Code} & \multirow{2}{=}{\centering Model Weights}  \\

        &&&&&&&&& \\
        
        \midrule
        
        MIMIC-Extract \cite{wang2020mimic} & MIMIC-III  & \checkmark & -- & 34k & 5 & -- & \checkmark & \checkmark & --  \\
        Purushotham 2018 \cite{purushotham2018benchmarking} & MIMIC-III & \checkmark & -- & 35k & 3 & -- & \checkmark & \checkmark & -- \\
        Harutyunyan 2019 \cite{harutyunyan2019multitask} & MIMIC-III & \checkmark & -- & 33k & 4 & -- & \checkmark & \checkmark & -- \\
        Gupta 2022 \cite{gupta2022extensive} & MIMIC-IV & \checkmark & * & 257k & 4 & -- & \checkmark & \checkmark & -- \\
        COP-E-CAT \cite{mandyam2021cop} & MIMIC-IV & \checkmark & * & 257k & 4 & -- & \checkmark & \checkmark & -- \\
        Xie 2022 \cite{xie2022benchmarking} & MIMIC-IV & \checkmark & * & 216k & 3 & -- & \checkmark & \checkmark & -- \\
        eICU \cite{sheikhalishahi2020benchmarking} & eICU & \checkmark & -- & 73k & 4 & -- & \checkmark & \checkmark & -- \\
        EHR PT \cite{mcdermott2021comprehensive} & MIMIC-III / eICU & \checkmark & -- & 86k & 11 & \checkmark & \checkmark & \checkmark & -- \\
        FIDDLE \cite{tang2020democratizing} & MIMIC-III / eICU & \checkmark & -- & 157k & 3 & -- & \checkmark & \checkmark & -- \\
        HiRID-ICU \cite{yeche2021hirid} & HiRID & \checkmark & -- & 33k & 6 & -- & \checkmark & \checkmark & -- \\
        Solares 2020 \cite{solares2020deep} & CPRD & \checkmark & \checkmark & 4M & 2 & -- & -- & -- & -- \\

        \midrule
        {\sc \textbf{EHRSHOT}} & Stanford Medicine & \checkmark & \checkmark & 7k & 15 & \checkmark & \checkmark & \checkmark & \checkmark \\
        
        \bottomrule
    \end{tabular}
  \label{tab:prior_work} 
\end{table}

%
%

\section{Dataset}

We are releasing {\sc EHRSHOT} (pronounced "earshot"), an EHR benchmark for few-shot evaluation of foundation models. {\sc EHRSHOT} is a collection of 6,739 unique patients with canonical train/validation/test splits and corresponding labels for 15 classification tasks. We also provide canonical $k$-shot samples for each few-shot evaluation task. Unlike prior EHR benchmarks focused on task-specific supervised models \cite{mcdermott2021comprehensive} for specific episodes of care, e.g. admission to the ICU \cite{harutyunyan2019multitask, pollard2018eicu}, our benchmark is designed for evaluating pretrained FMs on a broad range of tasks using the depth of information that a health system would typically possess for its patients. {\sc EHRSHOT} is provided as a set of CSV files. It is essentially a lightweight serialization of the OMOP-CDM format. Please see Section \ref{section_data_format} in the Appendix for additional details on the dataset format.

{\sc EHRSHOT} contains a total of 41.6 million coded observations (e.g. diagnoses, procedures, medications, lab results, etc.) and 921,499 unique visits across 6,739 patients. We exclude all patients less than 19 years of age or greater than 88 years of age. We also exclude patients with less than 10 total clinical events in their record. We include statistics of EHRSHOT's cohort demographics in Table \ref{tab:patient_summary_statistics} and Appendix Table \ref{tab:demographics}, and histograms of patient characteristics in Appendix Figure \ref{fig:timeline_histograms}.

\begin{table}[t]
    \small
    \centering 
    \caption{Summary statistics on the number of events, visits, and length of patient timelines in {\sc EHRSHOT}.}
    \vspace{0.2cm}

    \begin{tabular}{llllll}
        \toprule
        \textbf{Attribute} & & \textbf{Train} & \textbf{Val} & \textbf{Test} & \textbf{All Splits}\\
        \midrule
        \multirow{3}{*}{\textbf{Number of Events}} 
        & Min  & 10 & 10 & 10 & 10 \\
        & Mean & 5942 & 6758 & 5826 & 6174 \\
        & Max  & 113466 & 199913 & 129704 & 199913 \\~\\
        \multirow{3}{*}{\textbf{Number of Visits}} 
        & Min  & 0 & 0 & 0 & 0 \\
        & Mean & 127 & 147 & 134 & 136\\
        & Max  & 2099 & 2397 & 2023 & 2397 \\~\\
        \multirow{3}{*}{\textbf{Timeline Length (yrs)}} 
        & Min  & 19 & 19 & 19 & 19 \\
        & Mean & 59 & 59 & 58 & 59 \\
        & Max  & 88 & 88 & 88 & 88 \\
        \bottomrule
    \end{tabular}
    \label{tab:patient_summary_statistics} 
\end{table}

\subsection{Data Source} 
\label{section_data_source}

We sourced the data for our benchmark from the Stanford Medicine Research Data Repository (STARR) \cite{datta2020new}, which contains EHR data from both Stanford Health Care (primarily adult care) and Lucile Packard Children's Hospital (primarily pediatric care). The source dataset is structured according to the Observational Medical Outcomes Partnership Common Data Model (OMOP-CDM) \cite{hripcsak2015observational} and comprises a total of 3.67M unique patients from 1990 to February 8th, 2023 \cite{datta2020new}. Of these patients, 2.57M (70\%) are used for training and 0.55M (15\%) for validation of the foundation model that we release, CLMBR-T-base, the details of which we discuss in Section \ref{section_baselines}. All data that we work with is deidentified, and hence, our study did not require Institutional Review Board approval \cite{datta2020new}. 

This source database contains demographics (e.g. age, sex, race), diagnoses, procedures, laboratory results, medication prescriptions, and other coded clinical observations, which we preserve. While the source database also contains clinical notes, we remove these in our released benchmark. We describe how we selected our patient cohort from this source dataset in the Appendix in Section \ref{section_cohort_generation}. We apply a few additional transformations on top of those described in \cite{datta2020new} to prevent data leakage and fix timestamp issues, which are detailed in Section \ref{section_data_preprocessing} in the Appendix. 

For our data preprocessing pipeline, we use the \textbf{Framework for Electronic Medical Records (FEMR)} library, which we developed in parallel to this work. FEMR is a Python library that supports the ingestion of multiple EHR data formats (e.g. OMOP, MIMIC, etc.) and provides a unified interface for building machine learning models on top of such data at scale. The full codebase is available on Github here: \href{https://github.com/som-shahlab/femr/}{https://github.com/som-shahlab/femr/}.

Additionally, all of the code used to generate the dataset for {\sc EHRSHOT} can be found here: \href{https://github.com/som-shahlab/ehrshot-benchmark}{https://github.com/som-shahlab/ehrshot-benchmark}.

\begin{figure}[t]
    \centering    
    \includegraphics[scale=0.4]{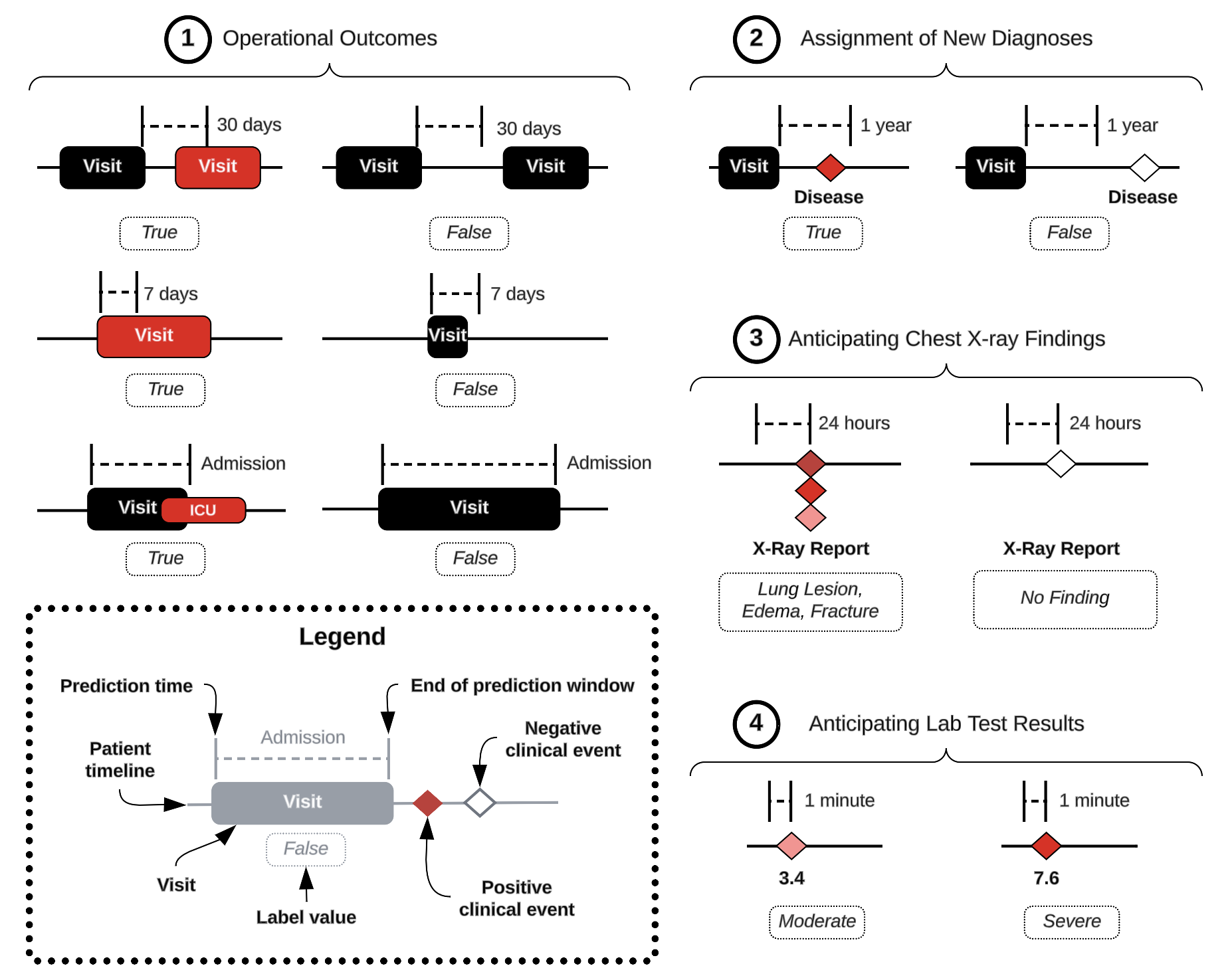}
    \caption{Summary of Benchmark Tasks. Each subfigure contains one of the 4 types of predictive classification tasks included in our benchmark: (1) \textit{Operational Outcomes} (binary), (2) \textit{Assignment of New Diagnoses} (binary), (3) \textit{Anticipating Chest X-ray Findings} (multilabel), (4) \textit{Anticipating Lab Test Results} (multiclass). Each \textbf{black line} represents a patient timeline. The \fbox{black boxes} represent how each timeline would be labeled for each task at a specific prediction time. The leftmost edge of the \dotuline{dotted lines} above each timeline is the prediction time, and the rightmost edge is the end of the time horizon for that task. Note that each \textit{Operational Outcome} task (1) has a different prediction window (30 days, 7 days, duration of admission), while the other three task categories (2, 3, 4) all have uniform prediction windows across their subtasks.}
    \label{fig:figure_2}
\end{figure}

\subsection{Tasks} 

We define 15 tasks as part of our benchmark, as listed in Table \ref{tab:task_demographics}. We selected these tasks based on clinician input as well as alignment with prior benchmarks \cite{harutyunyan2019multitask, guo2023ehr}. The tasks that we consider can be broadly grouped into the following 4 categories: (1) Operational Outcomes, (2) Anticipating Lab Test Values, (3) Assignment of New Diagnoses, (4) Anticipating Chest X-ray Findings. 

All tasks are classification tasks. We include a total of nine binary classification tasks (\textit{Operational Outcomes} and \textit{Assignment of New Diagnoses}), five 5-way multiclass tasks (\textit{Anticipating Lab Test Values}), and one 14-way multilabel task (\textit{Anticipating Chest X-ray Findings}). The size of each task's subcohort, as well as the prevalence of positive labels, is detailed in Table \ref{tab:task_demographics}. For example, there are 552 positive labels within the test cohort for the Long Length of Stay task, while there are 2,195 total labels, meaning there are 1,643 negative labels. As there are only 1,238 unique patients in this task's test cohort, some patients have multiple labels assigned to them. 

In the Appendix, we define the precise prediction windows for each task in  Table \ref{tab:task_definitions} and the definition of each task in Section \ref{section_task_definitions}. We also provide a visualization of our 4 task categories in Figure \ref{fig:figure_2}.

\begin{table}[t]
    \centering 
    \caption{Task Demographics. The number of unique patients and total labels for each task. A single patient may have multiple labels for one task (e.g. a patient with multiple anemia lab results). We show the prevalence of positive patients/labels in parenthesis. For the multiclass lab test tasks, we define a positive label as any non-normal result. For the multilabel chest X-ray task, we define a positive label as a report with at least one finding.}

    \vspace{0.2cm}
    \scriptsize

    \begin{tabular}{l ll l ll l ll}
    
        \toprule
        
        \multirow{4}{*}{\textbf{Task Name}} & \multicolumn{2}{c}{\textbf{Train}} && \multicolumn{2}{c}{\textbf{Val}} && \multicolumn{2}{c}{\textbf{Test}}  \\ 
        
        \cline{2-3} \cline{5-6} \cline{8-9}\\
        
        & \# Patients & \# Labels && \# Patients & \# Labels && \# Patients & \# Labels \\
        & (\# Positive) & (\# Positive) &&  (\# Positive) & (\# Positive) &&  (\# Positive) & (\# Positive) \\
        
        \midrule
        
        \multicolumn{2}{l}{\textbf{Operational Outcomes}}\\
        
            Long Length of Stay & 1377 (464) & 2569 (681) & & 1240 (395) & 2231 (534) & & 1238 (412) & 2195 (552) \\
            30-day Readmission & 1337 (164) & 2609 (370) & & 1192 (159) & 2207 (281) & & 1190 (151) & 2189 (260) \\
            ICU Transfer & 1306 (107) & 2402 (113) & & 1157 (84) & 2052 (92) & & 1154 (75) & 2037 (85) \\~\\

        \multicolumn{2}{l}{\textbf{Anticipating Lab Test Results}}\\
        
            Thrombocytopenia & 2084 (870) & 68776 (9774) & & 1981 (774) & 54504 (6962) & & 1998 (818) & 56338 (7960) \\
            Hyperkalemia & 2038 (383) & 76349 (1215) & & 1935 (348) & 60168 (886) & & 1958 (339) & 63653 (948) \\
            Hypoglycemia & 2054 (422) & 122108 (1065) & & 1950 (362) & 95488 (858) & & 1970 (356) & 100568 (783) \\
            Hyponatremia & 2035 (1288) & 81336 (20181) & & 1930 (1165) & 64473 (14674) & & 1956 (1212) & 67028 (16003) \\
            Anemia & 2092 (1251) & 70501 (9544) & & 1992 (1122) & 56224 (7445) & & 2002 (1151) & 58155 (7636) \\~\\

        \multicolumn{2}{l}{\textbf{Assignment of New Diagnoses}}\\
        
            Hypertension & 793 (130) & 1260 (184) & & 784 (130) & 1250 (177) & & 758 (130) & 1261 (160) \\
            Hyperlipidemia & 923 (137) & 1684 (205) & & 863 (140) & 1441 (189) & & 864 (133) & 1317 (172) \\
            Pancreatic Cancer & 1376 (128) & 2576 (155) & & 1242 (46) & 2215 (53) & & 1246 (40) & 2220 (56) \\
            Celiac & 1392 (48) & 2623 (62) & & 1252 (8) & 2284 (11) & & 1255 (13) & 2222 (21) \\
            Lupus & 1377 (79) & 2570 (104) & & 1239 (24) & 2226 (33) & & 1249 (19) & 2243 (20) \\
            Acute MI & 1365 (130) & 2534 (175) & & 1234 (112) & 2177 (146) & & 1235 (115) & 2127 (144) \\~\\
        
        \multicolumn{2}{l}{\textbf{Anticipating Chest X-ray Findings}}\\
            Chest X-Ray Findings & 251 (237) & 7481 (4771) & & 395 (378) & 9366 (6032) & & 399 (381) & 9428 (6400) \\

        \bottomrule
    \end{tabular}
    
    \label{tab:task_demographics} 

\end{table}

%
%

\section{Baseline Models}
\label{section_baselines}

We measure the performance of two baseline models on our dataset: (1) a gradient boosting machine (GBM) that uses count-based featurizations of patients to make predictions, (2) an autoregressive language model ("CLMBR-T-base") that ingests medical codes as tokens and was pretrained on the full longitudinal structured EHRs of 2.57M patients from our source institution \cite{steinberg2021language, guo2023ehr}. 

We chose these two models as our baselines for several reasons. First, language modeling has achieved state-of-the-art results on clinical prediction tasks \cite{steinberg2021language, rasmy2021med, munoz2022sehr, poulain2022few, li2020behrt}, while count-based featurization remains a simple but competitive baseline \cite{rajkomar2018scalable, rasmy2021med, steinberg2021language}. Second, most prior FMs trained on structured EHR data have not had their model weights published, and were developed and tested exclusively on nonstandard data formats like MIMIC-III \cite{wornow2023shaky}. This makes it nearly impossible to conduct a fair comparison of prior models, which often requires re-implementation or significant modification to work across datasets \cite{hur2021unifying}. This is one of the key challenges we are attempting to solve with {\sc EHRSHOT}. We pre-train our own FM from scratch to have full control over its training, and publish its model weights so the community can reproduce and build upon our results.

\textbf{Count-based Features}. Count-based featurization is a well-established baseline for EHR tasks, valued for its simplicity and effectiveness \cite{rajkomar2018scalable}. The fundamental idea involves converting each patient's timeline into a count vector, where each element contains the number of occurrences of a specific medical concept prior to the prediction time of a task. These patient vectors are combined into a count matrix, which is high-dimensional and sparse. We use a technique called \textit{ontology expansion} to increase the density of representation and improve the accuracy of code coverage by acknowledging the parent/child hierarchical relationships between medical concepts \cite{choi2017gram}. After generating our ontology-expanded count matrix, we train a gradient boosting machine (GBM) model on the {\sc EHRSHOT} train split, and tune hyperparameters on the validation split. We use the LightGBM implementation \cite{ke2017lightgbm}. We also evaluate a Logistic Regression and Random Forest model as baselines. Their results can be seen in Appendix in Figures \ref{fig:agg_auroc_all} and \ref{fig:agg_auprc_all}. For clarity, we exclude them from the following analyses, as they perform roughly at par with the count-based GBM model. 

\textbf{Clinical Language-Model-Based Representations using Transformers (CLMBR-T-base)}. CLMBR-T-base is an autoregressive model designed to predict the next medical code in a patient's timeline given previous codes. This objective enables it to learn robust global patterns for clinical prediction tasks. It is based on the CLMBR model originally developed in \cite{steinberg2021language}, but following \cite{guo2023ehr} we substitute a transformer in place of a GRU as its base model. Our model employs causally masked local attention. This ensures forward-only flow of information which is vital for prediction tasks, and is in contrast to BERT-based models which are bidirectional in nature \cite{steinberg2021language}. Note that our model does not process clinical text, only structured information. Our model has 141M trainable parameters, a hidden dimension of 768, and a next code prediction objective. This provides our version of CLMBR-T-base with minute-level resolution rather than the day-level aggregation of the original model formulation \cite{steinberg2021language}. We leave training larger versions of CLMBR to future work.

More details about our baseline models can be found in the Appendix in Section \ref{section_results_details}.

\section{Results}
\label{section_results}

We evaluate each baseline model in a few-shot setting. For each of the 15 benchmark tasks, we steadily increase the number of examples $k$ that each model sees from $k = 1$ to the full training dataset, and record the model's AUROC and AUPRC at each $k$.

More precisely, we define "$k$-shot evaluation" of a model $M$ on a specific task $T$ as follows. We train $M$ on $k$ positive examples and $k$ negative examples sampled from $T$'s training split. We then select an additional $k$ positive examples and $k$ negative examples from $T$'s validation split, and use these validation examples to select the best hyperparameters for $M$ for task $T$. Finally, we evaluate the AUROC and AUPRC of the best performing version of $M$ on $T$'s entire held-out test split. For tasks where the total number of unique positive examples is less than $k$, we include all positive examples in our training set, and randomly resample positive examples until the total number of training examples seen by the model is $k$. We consider values of $k \in \{ 1, 2, 4, 8, 12, 16, 24, 32, 48, 64, 128 \}$ for all tasks (with the exception of Celiac, for which we limit $k \le 64$ as there are only 62 positive training labels).

For the count-based GBM, these few-shot examples are the only training examples seen by the model. For the pretrained CLMBR-T-base model, we use these few-shot examples to fine-tune a logistic regression head appended to the top of the model, while keeping the weights of the pretrained CLMBR-T-base model frozen. Pretraining the CLMBR-T-base model took roughly 4 days on a single Nvidia V100 hosted in an on-premise compute cluster.

The AUROC of each model across all 4 task categories is presented in Figure \ref{fig:agg_auroc}. In the Appendix, we show this grouping for AUPRC in Figure \ref{fig:agg_auprc}. We also break down each individual task's AUROC in Figure \ref{fig:separate_auroc} and AUPRC in Figure \ref{fig:separate_auprc} of the Appendix. We also include results for additional baselines in the Appendix in Figures \ref{fig:agg_auroc_all} and \ref{fig:agg_auprc_all}. The \textbf{bolded lines} are the Macro-AUC for each model within a task category, averaged across all subtasks at each $k$. We include the performance of each model trained on the entire EHRSHOT training split on the far right of every plot as \textit{All}. 

As shown in Figure \ref{fig:agg_auroc}, the pretrained foundation model CLMBR-T-base (\textcolor{blue}{blue}) outperforms the count-based GBM (\textcolor{red}{red}) across all aggregated task categories for $k \le 64$. This demonstrates the benefits of pretraining in few-shot settings, as the model can leverage patterns learned across millions of patients to derive more accurate representations out-of-the-box than a model trained from scratch. CLMBR-T-base outperforms the count-based GBM across all $k$ on the \textit{Operational Outcomes} and the majority of \textit{Anticipating Lab Test Results} and \textit{Anticipating Chest X-ray Findings} tasks. For these three task groups, the advantage of CLMBR-T-base seems most pronounced at intermediate levels of $k$ between 8 and 128. At extremely low $k$ (i.e. $k = 1$), both models struggle to learn anything, while as $k$ increases the advantage of the pretrained model tends to shrink, a trend noted elsewhere \cite{mcdermott2021comprehensive}. This is most visible in the far-right of the plot at the \textit{All} marker, which represents the performance of each model when trained on the full EHRSHOT training dataset.

\begin{figure}[t]
    \centering
    \includegraphics[width=\textwidth]{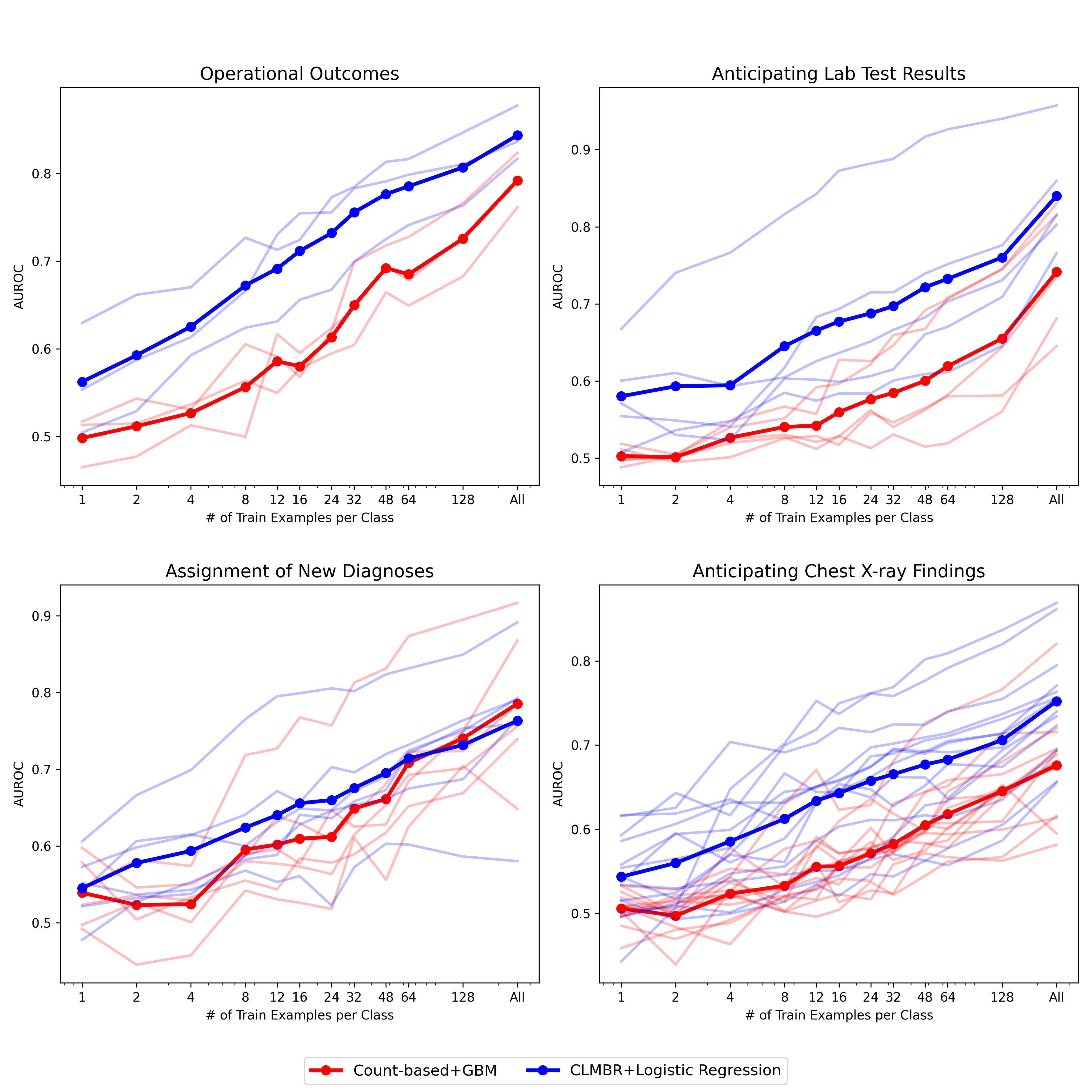}
    \caption{Aggregated AUROC across all subtasks within each of the 4 task categories for $k \in \{1, 2, 4, 8, 12, 16, 24, 32, 48, 64, 128 \}$ shots. We show performance on the full training set as \textit{All}. The \textbf{bolded lines} are the Macro-AUROC for each model, averaged across all subtasks within a task category for each value of $k$. The blurred lines are the average AUROC across 5 replicates for each subtask within a task category. CLMBR-T-base (\textcolor{blue}{blue}) consistently outperforms the count-based GBM (\textcolor{red}{red}) at $k \le 64$, but lags in higher label settings for the \textit{Assignment of New Diagnoses} tasks.}
    \label{fig:agg_auroc}
\end{figure}

In fact, the count-based GBM exceeds the performance of CLMBR-T-base on the \textit{Assignment of New Diagnoses} tasks at $k > 64$. This suggests that the advantage of pretraining comes primarily from improved initialization of patient representations, and that the largest gains are achieved in the most data poor regimes. 

There are several possible reasons for CLMBR-T-base's underperformance at higher values of $k$ for the \textit{Assignment of New Diagnoses} tasks. First, the CLMBR-T-base model's training objective is next code prediction, which makes it ill-suited for predictive tasks with long time horizons (which for these tasks is 1 year). Second, if a simple tree-based model exists for a task (i.e. a few medical concepts tightly correlate with a diagnosis), then it may be more difficult for a pretrained model to coerce patient representations learned over millions of patients to that specific task than training a model from scratch with enough data to learn those distinctive signals. We believe that this reversal in model rankings demonstrates a key strength of {\sc EHRSHOT} -- namely, the diversity of its predictive tasks can help identify opportunities for improving pretraining and few-shot strategies.

We release all of our model weights, evaluation tasks, and data processing code to fully reproduce our results. To the best of our knowledge, the release of our pretrained CLMBR-T-base model is one of the first examples of such a clinical FM having its pretrained weights made publicly available \cite{wornow2023shaky}.

%
%

\section{Discussion}
\label{section_discussion}

We believe that {\sc EHRSHOT} represents a useful contribution to the ML community by enabling more reproducible healthcare ML research. The release of our pretrained CLMBR-T-base model's weights will allow the community to replicate and build upon our work. Our results identify opportunities for improving pretrained models in few-shot settings.

Acquiring labeled EHR data is expensive and time-consuming. Additionally, certain rare conditions may only be present in a small cohort of patients out of millions within a health system \cite{poulain2022few}. Thus, model performance in low-label settings is of paramount importance in healthcare. As our results in Section \ref{section_results} demonstrate, pretrained FMs can yield large performance gains in few-shot settings. While we acknowledge that the tasks themselves may not be the most clinically meaningful, we believe that {\sc EHRSHOT} offers a valuable contribution by providing a reproducible and rigorous point of comparison for different technical approaches to developing clinical FMs.

\textbf{Limitations}. There are several limitations to this work. First, we only release structured data -- i.e. we do not publish any of the clinical text or images associated with our patients. While many datasets for medical images exist \cite{ccalli2021deep}, publishing clinical text remains a challenge \cite{spasic2020clinical}. Second, we only consider one type of foundation model (CLMBR-T-base) for our experiments \cite{steinberg2021language}. We look forward to seeing the additional foundation models that the community applies to our benchmark. Third, we release a very small cohort of patients (<1\%) from our source EHR database, and specifically select these patients for the tasks that we define. Releasing our full pretraining dataset would be infeasible from a governance and effort perspective. Thus, while necessary in order to publish our EHR dataset and still broader than existing ICU-specific datasets, our cohort selection process limits the types of questions we can answer and does not reflect the full diversity of medical data. Fourth, as we only were able to evaluate our pretrained FM on Stanford Medicine data, it is unclear how well our pretrained model will perform at other institutions. We anticipate there will be some drop in performance, but the extent is unclear. Fifth, several of our tasks are "low label" in the most extreme sense -- for example, the Celiac task only has 13 positive patients in its test set. This makes obtaining low variance estimates of model performance difficult. We aim to mitigate this by adding additional patients to our benchmark in future releases.

\textbf{Societal Implications}. We believe that the release of this dataset can help spur positive innovations for improving clinical care with ML. However, we recognize that there are patient privacy concerns anytime EHR data is released. We believe we sufficiently mitigate this risk through the rigorous deidentification process on which our data is subjected \cite{datta2020new}. Additionally, we gate access to the dataset through a research data use agreement. Another concern is that models trained on biased data will reflect those biases \cite{gianfrancesco2018potential}. Thus, the pretrained FM that we release may propagate biases in care delivery or outcomes present in our source EHR database \cite{gianfrancesco2018potential}. However, we hope that by encouraging the full release of models, we can help the community better identify and mitigate these issues \cite{norori2021addressing}.

\section{Conclusion}
\label{section_conclusion}

We publish {\sc EHRSHOT}, a benchmark containing the structured data of 6,739 patients' full longitudinal medical timelines specifically geared towards few-shot evaluation of foundation models for clinical data. Unlike most prior work, {\sc EHRSHOT} contains longitudinal health data rather than a single department (e.g. ICU). We define a set of 15 tasks ranging from well-studied outcomes like 30-day readmission to lesser explored settings such as anticipating abnormal lab values. Finally, we release the weights of a foundation model pretrained on over 2.57M patient timelines and publish the code needed to replicate our results. We hope that this work represents a first step towards moving the field of ML for healthcare towards more reproducible and open model development.

\begin{ack}
We thank the Stanford AIMI Center and Stanford Medicine Research IT for their assistance in publishing this dataset. This work was supported in part by the Mark and Debra Leslie Endowment for AI in Healthcare, the Clinical Excellence Research Center at Stanford Medicine, and Technology and Digital Solutions at Stanford Healthcare. MW is supported by an NSF Graduate Research Fellowship. JF was supported in part by a Stanford AIMI-HAI Partnership Grant. 
\end{ack}

\newpage

\section*{}
{
\small
\bibliographystyle{plain}
\bibliography{bibliography.bib}

\begin{thebibliography}{10}

\bibitem{opportunities_and_risks}
Rishi Bommasani, Drew~A. Hudson, Ehsan Adeli, Russ~B. Altman, Simran Arora,
  Sydney von Arx, Michael~S. Bernstein, Jeannette Bohg, Antoine Bosselut, Emma
  Brunskill, Erik Brynjolfsson, Shyamal Buch, Dallas Card, Rodrigo Castellon,
  Niladri~S. Chatterji, Annie~S. Chen, Kathleen Creel, Jared~Quincy Davis,
  Dorottya Demszky, Chris Donahue, Moussa Doumbouya, Esin Durmus, Stefano
  Ermon, John Etchemendy, Kawin Ethayarajh, Li~Fei{-}Fei, Chelsea Finn, Trevor
  Gale, Lauren Gillespie, Karan Goel, Noah~D. Goodman, Shelby Grossman, Neel
  Guha, Tatsunori Hashimoto, Peter Henderson, John Hewitt, Daniel~E. Ho, Jenny
  Hong, Kyle Hsu, Jing Huang, Thomas Icard, Saahil Jain, Dan Jurafsky,
  Pratyusha Kalluri, Siddharth Karamcheti, Geoff Keeling, Fereshte Khani, Omar
  Khattab, Pang~Wei Koh, Mark~S. Krass, Ranjay Krishna, Rohith Kuditipudi, and
  et~al.
\newblock On the opportunities and risks of foundation models.
\newblock {\em CoRR}, abs/2108.07258, 2021.

\bibitem{bycroft2018uk}
Clare Bycroft, Colin Freeman, Desislava Petkova, Gavin Band, Lloyd~T Elliott,
  Kevin Sharp, Allan Motyer, Damjan Vukcevic, Olivier Delaneau, Jared
  O’Connell, et~al.
\newblock The uk biobank resource with deep phenotyping and genomic data.
\newblock {\em Nature}, 562(7726):203--209, 2018.

\bibitem{ccalli2021deep}
Erdi {\c{C}}all{\i}, Ecem Sogancioglu, Bram van Ginneken, Kicky~G van Leeuwen,
  and Keelin Murphy.
\newblock Deep learning for chest x-ray analysis: A survey.
\newblock {\em Medical Image Analysis}, 72:102125, 2021.

\bibitem{choi2017gram}
Edward Choi, Mohammad~Taha Bahadori, Le~Song, Walter~F Stewart, and Jimeng Sun.
\newblock Gram: graph-based attention model for healthcare representation
  learning.
\newblock In {\em Proceedings of the 23rd ACM SIGKDD international conference
  on knowledge discovery and data mining}, pages 787--795, 2017.

\bibitem{datta2020new}
Somalee Datta, Jose Posada, Garrick Olson, Wencheng Li, Ciaran O'Reilly, Deepa
  Balraj, Joseph Mesterhazy, Joseph Pallas, Priyamvada Desai, and Nigam Shah.
\newblock A new paradigm for accelerating clinical data science at stanford
  medicine.
\newblock {\em arXiv preprint arXiv:2003.10534}, 2020.

\bibitem{faltys10hirid}
M~Faltys, M~Zimmermann, X~Lyu, M~H{\"u}ser, S~Hyland, G~R{\"a}tsch, and T~Merz.
\newblock Hirid, a high time-resolution icu dataset (version 1.0). physionet.
  2020.
\newblock {\em DOI: https://doi. org/10.13026/hz5m-md48}, 2020.

\bibitem{gianfrancesco2018potential}
Milena~A Gianfrancesco, Suzanne Tamang, Jinoos Yazdany, and Gabriela Schmajuk.
\newblock Potential biases in machine learning algorithms using electronic
  health record data.
\newblock {\em JAMA internal medicine}, 178(11):1544--1547, 2018.

\bibitem{guo2023ehr}
Lin~Lawrence Guo, Ethan Steinberg, Scott~Lanyon Fleming, Jose Posada, Joshua
  Lemmon, Stephen~R Pfohl, Nigam Shah, Jason Fries, and Lillian Sung.
\newblock Ehr foundation models improve robustness in the presence of temporal
  distribution shift.
\newblock {\em Scientific Reports}, 13(1):3767, 2023.

\bibitem{gupta2022extensive}
Mehak Gupta, Brennan Gallamoza, Nicolas Cutrona, Pranjal Dhakal, Raphael
  Poulain, and Rahmatollah Beheshti.
\newblock An extensive data processing pipeline for mimic-iv.
\newblock In {\em Machine Learning for Health}, pages 311--325. PMLR, 2022.

\bibitem{harutyunyan2019multitask}
Hrayr Harutyunyan, Hrant Khachatrian, David~C Kale, Greg Ver~Steeg, and Aram
  Galstyan.
\newblock Multitask learning and benchmarking with clinical time series data.
\newblock {\em Scientific data}, 6(1):96, 2019.

\bibitem{herrett2015data}
Emily Herrett, Arlene~M Gallagher, Krishnan Bhaskaran, Harriet Forbes, Rohini
  Mathur, Tjeerd Van~Staa, and Liam Smeeth.
\newblock Data resource profile: clinical practice research datalink (cprd).
\newblock {\em International journal of epidemiology}, 44(3):827--836, 2015.

\bibitem{hripcsak2015observational}
George Hripcsak, Jon~D Duke, Nigam~H Shah, Christian~G Reich, Vojtech Huser,
  Martijn~J Schuemie, Marc~A Suchard, Rae~Woong Park, Ian Chi~Kei Wong, Peter~R
  Rijnbeek, et~al.
\newblock Observational health data sciences and informatics (ohdsi):
  opportunities for observational researchers.
\newblock In {\em MEDINFO 2015: eHealth-enabled Health}, pages 574--578. IOS
  Press, 2015.

\bibitem{hur2021unifying}
Kyunghoon Hur, Jiyoung Lee, Jungwoo Oh, Wesley Price, Young-Hak Kim, and Edward
  Choi.
\newblock Unifying heterogenous electronic health records systems via
  text-based code embedding.
\newblock {\em arXiv preprint arXiv:2111.09098}, 2021.

\bibitem{irvin2019chexpert}
Jeremy Irvin, Pranav Rajpurkar, Michael Ko, Yifan Yu, Silviana Ciurea-Ilcus,
  Chris Chute, Henrik Marklund, Behzad Haghgoo, Robyn Ball, Katie Shpanskaya,
  et~al.
\newblock Chexpert: A large chest radiograph dataset with uncertainty labels
  and expert comparison.
\newblock In {\em Proceedings of the AAAI conference on artificial
  intelligence}, volume~33, pages 590--597, 2019.

\bibitem{jain2021radgraph}
Saahil Jain, Ashwin Agrawal, Adriel Saporta, Steven~QH Truong, Du~Nguyen Duong,
  Tan Bui, Pierre Chambon, Yuhao Zhang, Matthew~P Lungren, Andrew~Y Ng, et~al.
\newblock Radgraph: Extracting clinical entities and relations from radiology
  reports.
\newblock {\em arXiv preprint arXiv:2106.14463}, 2021.

\bibitem{johnson2023mimic}
Alistair~EW Johnson, Lucas Bulgarelli, Lu~Shen, Alvin Gayles, Ayad Shammout,
  Steven Horng, Tom~J Pollard, Benjamin Moody, Brian Gow, Li-wei~H Lehman,
  et~al.
\newblock Mimic-iv, a freely accessible electronic health record dataset.
\newblock {\em Scientific data}, 10(1):1, 2023.

\bibitem{johnson2016mimic}
Alistair~EW Johnson, Tom~J Pollard, Lu~Shen, Li-wei~H Lehman, Mengling Feng,
  Mohammad Ghassemi, Benjamin Moody, Peter Szolovits, Leo Anthony~Celi, and
  Roger~G Mark.
\newblock Mimic-iii, a freely accessible critical care database.
\newblock {\em Scientific data}, 3(1):1--9, 2016.

\bibitem{ke2017lightgbm}
Guolin Ke, Qi~Meng, Thomas Finley, Taifeng Wang, Wei Chen, Weidong Ma, Qiwei
  Ye, and Tie-Yan Liu.
\newblock Lightgbm: A highly efficient gradient boosting decision tree.
\newblock {\em Advances in neural information processing systems}, 30, 2017.

\bibitem{langenkamp2022open}
Max Langenkamp and Daniel~N Yue.
\newblock How open source machine learning software shapes ai.
\newblock In {\em Proceedings of the 2022 AAAI/ACM Conference on AI, Ethics,
  and Society}, pages 385--395, 2022.

\bibitem{li2020behrt}
Yikuan Li, Shishir Rao, Jos{\'e} Roberto~Ayala Solares, Abdelaali Hassaine,
  Rema Ramakrishnan, Dexter Canoy, Yajie Zhu, Kazem Rahimi, and Gholamreza
  Salimi-Khorshidi.
\newblock Behrt: transformer for electronic health records.
\newblock {\em Scientific reports}, 10(1):1--12, 2020.

\bibitem{mandyam2021cop}
Aishwarya Mandyam, Elizabeth~C Yoo, Jeff Soules, Krzysztof Laudanski, and
  Barbara~E Engelhardt.
\newblock Cop-e-cat: cleaning and organization pipeline for ehr computational
  and analytic tasks.
\newblock In {\em Proceedings of the 12th ACM Conference on Bioinformatics,
  Computational Biology, and Health Informatics}, pages 1--9, 2021.

\bibitem{mcdermott2021comprehensive}
Matthew McDermott, Bret Nestor, Evan Kim, Wancong Zhang, Anna Goldenberg, Peter
  Szolovits, and Marzyeh Ghassemi.
\newblock A comprehensive ehr timeseries pre-training benchmark.
\newblock In {\em Proceedings of the Conference on Health, Inference, and
  Learning}, pages 257--278, 2021.

\bibitem{mcdermott2021reproducibility}
Matthew~BA McDermott, Shirly Wang, Nikki Marinsek, Rajesh Ranganath, Luca
  Foschini, and Marzyeh Ghassemi.
\newblock Reproducibility in machine learning for health research: Still a ways
  to go.
\newblock {\em Science Translational Medicine}, 13(586):eabb1655, 2021.

\bibitem{moor2023foundation}
Michael Moor, Oishi Banerjee, Zahra Shakeri~Hossein Abad, Harlan~M Krumholz,
  Jure Leskovec, Eric~J Topol, and Pranav Rajpurkar.
\newblock Foundation models for generalist medical artificial intelligence.
\newblock {\em Nature}, 616(7956):259--265, 2023.

\bibitem{munoz2022sehr}
Anna Munoz-Farre, Harry Rose, and Sera~Aylin Cakiroglu.
\newblock sehr-ce: Language modelling of structured ehr data for efficient and
  generalizable patient cohort expansion.
\newblock {\em arXiv preprint arXiv:2211.17121}, 2022.

\bibitem{norori2021addressing}
Natalia Norori, Qiyang Hu, Florence~Marcelle Aellen, Francesca~Dalia Faraci,
  and Athina Tzovara.
\newblock Addressing bias in big data and ai for health care: A call for open
  science.
\newblock {\em Patterns}, 2(10):100347, 2021.

\bibitem{piwowar2013data}
Heather~A Piwowar and Todd~J Vision.
\newblock Data reuse and the open data citation advantage.
\newblock {\em PeerJ}, 1:e175, 2013.

\bibitem{pollard2018eicu}
Tom~J Pollard, Alistair~EW Johnson, Jesse~D Raffa, Leo~A Celi, Roger~G Mark,
  and Omar Badawi.
\newblock The eicu collaborative research database, a freely available
  multi-center database for critical care research.
\newblock {\em Scientific data}, 5(1):1--13, 2018.

\bibitem{poulain2022few}
Raphael Poulain, Mehak Gupta, and Rahmatollah Beheshti.
\newblock Few-shot learning with semi-supervised transformers for electronic
  health records.
\newblock In Zachary Lipton, Rajesh Ranganath, Mark Sendak, Michael Sjoding,
  and Serena Yeung, editors, {\em Proceedings of the 7th Machine Learning for
  Healthcare Conference}, volume 182 of {\em Proceedings of Machine Learning
  Research}, pages 853--873. PMLR, 05--06 Aug 2022.

\bibitem{purushotham2018benchmarking}
Sanjay Purushotham, Chuizheng Meng, Zhengping Che, and Yan Liu.
\newblock Benchmarking deep learning models on large healthcare datasets.
\newblock {\em Journal of biomedical informatics}, 83:112--134, 2018.

\bibitem{qiu2023large}
Jianing Qiu, Lin Li, Jiankai Sun, Jiachuan Peng, Peilun Shi, Ruiyang Zhang,
  Yinzhao Dong, Kyle Lam, Frank P-W Lo, Bo~Xiao, et~al.
\newblock Large ai models in health informatics: Applications, challenges, and
  the future.
\newblock {\em arXiv preprint arXiv:2303.11568}, 2023.

\bibitem{rajkomar2018scalable}
Alvin Rajkomar, Eyal Oren, Kai Chen, Andrew~M Dai, Nissan Hajaj, Michaela
  Hardt, Peter~J Liu, Xiaobing Liu, Jake Marcus, Mimi Sun, et~al.
\newblock Scalable and accurate deep learning with electronic health records.
\newblock {\em NPJ digital medicine}, 1(1):18, 2018.

\bibitem{rasmy2021med}
Laila Rasmy, Yang Xiang, Ziqian Xie, Cui Tao, and Degui Zhi.
\newblock Med-bert: pretrained contextualized embeddings on large-scale
  structured electronic health records for disease prediction.
\newblock {\em NPJ digital medicine}, 4(1):86, 2021.

\bibitem{russakovsky2015imagenet}
Olga Russakovsky, Jia Deng, Hao Su, Jonathan Krause, Sanjeev Satheesh, Sean Ma,
  Zhiheng Huang, Andrej Karpathy, Aditya Khosla, Michael Bernstein, et~al.
\newblock Imagenet large scale visual recognition challenge.
\newblock {\em International journal of computer vision}, 115:211--252, 2015.

\bibitem{omop-cdm}
Observational Health~Data Sciences and Informatics.
\newblock The book of ohdsi, Jan 2021.

\bibitem{sevilla2022compute}
Jaime Sevilla, Lennart Heim, Anson Ho, Tamay Besiroglu, Marius Hobbhahn, and
  Pablo Villalobos.
\newblock Compute trends across three eras of machine learning.
\newblock In {\em 2022 International Joint Conference on Neural Networks
  (IJCNN)}, pages 1--8. IEEE, 2022.

\bibitem{sheikhalishahi2020benchmarking}
Seyedmostafa Sheikhalishahi, Vevake Balaraman, and Venet Osmani.
\newblock Benchmarking machine learning models on multi-centre eicu critical
  care dataset.
\newblock {\em Plos one}, 15(7):e0235424, 2020.

\bibitem{sohn2023reproducibility}
Emily Sohn.
\newblock The reproducibility issues that haunt health-care ai.
\newblock {\em Nature}, 613:402--403, 01 2023.

\bibitem{solares2020deep}
Jose Roberto~Ayala Solares, Francesca Elisa~Diletta Raimondi, Yajie Zhu,
  Fatemeh Rahimian, Dexter Canoy, Jenny Tran, Ana Catarina~Pinho Gomes, Amir~H
  Payberah, Mariagrazia Zottoli, Milad Nazarzadeh, et~al.
\newblock Deep learning for electronic health records: A comparative review of
  multiple deep neural architectures.
\newblock {\em Journal of biomedical informatics}, 101:103337, 2020.

\bibitem{sonnenburg2007need}
Soren Sonnenburg, Mikio~L Braun, Cheng~Soon Ong, Samy Bengio, Leon Bottou,
  Geoffrey Holmes, Yann LeCunn, Klaus-Robert Muller, Fernando Pereira,
  Carl~Edward Rasmussen, et~al.
\newblock The need for open source software in machine learning.
\newblock {\em Journal of Machine Learning Research}, 8(81):2443--2466, 2007.

\bibitem{spasic2020clinical}
Irena Spasic, Goran Nenadic, et~al.
\newblock Clinical text data in machine learning: systematic review.
\newblock {\em JMIR medical informatics}, 8(3):e17984, 2020.

\bibitem{steinberg2021language}
Ethan Steinberg, Ken Jung, Jason~A Fries, Conor~K Corbin, Stephen~R Pfohl, and
  Nigam~H Shah.
\newblock Language models are an effective representation learning technique
  for electronic health record data.
\newblock {\em Journal of biomedical informatics}, 113:103637, 2021.

\bibitem{su2021roformer}
Jianlin Su, Yu~Lu, Shengfeng Pan, Ahmed Murtadha, Bo~Wen, and Yunfeng Liu.
\newblock Roformer: Enhanced transformer with rotary position embedding.
\newblock {\em arXiv preprint arXiv:2104.09864}, 2021.

\bibitem{tang2020democratizing}
Shengpu Tang, Parmida Davarmanesh, Yanmeng Song, Danai Koutra, Michael~W
  Sjoding, and Jenna Wiens.
\newblock Democratizing ehr analyses with fiddle: a flexible data-driven
  preprocessing pipeline for structured clinical data.
\newblock {\em Journal of the American Medical Informatics Association},
  27(12):1921--1934, 2020.

\bibitem{thoral2021sharing}
Patrick~J Thoral, Jan~M Peppink, Ronald~H Driessen, Eric~JG Sijbrands, Erwin~JO
  Kompanje, Lewis Kaplan, Heatherlee Bailey, Jozef Kesecioglu, Maurizio
  Cecconi, Matthew Churpek, et~al.
\newblock Sharing icu patient data responsibly under the society of critical
  care medicine/european society of intensive care medicine joint data science
  collaboration: the amsterdam university medical centers database
  (amsterdamumcdb) example.
\newblock {\em Critical care medicine}, 49(6):e563, 2021.

\bibitem{toscher2009bigchaos}
Andreas T{\"o}scher, Michael Jahrer, and Robert~M Bell.
\newblock The bigchaos solution to the netflix grand prize.
\newblock {\em Netflix prize documentation}, pages 1--52, 2009.

\bibitem{van2023yet}
Robin van~de Water, Hendrik Schmidt, Paul Elbers, Patrick Thoral, Bert Arnrich,
  and Patrick Rockenschaub.
\newblock Yet another icu benchmark: A flexible multi-center framework for
  clinical ml.
\newblock {\em arXiv preprint arXiv:2306.05109}, 2023.

\bibitem{wang2020mimic}
Shirly Wang, Matthew~BA McDermott, Geeticka Chauhan, Marzyeh Ghassemi,
  Michael~C Hughes, and Tristan Naumann.
\newblock Mimic-extract: A data extraction, preprocessing, and representation
  pipeline for mimic-iii.
\newblock In {\em Proceedings of the ACM conference on health, inference, and
  learning}, pages 222--235, 2020.

\bibitem{wornow2023shaky}
Michael Wornow, Yizhe Xu, Rahul Thapa, Birju Patel, Ethan Steinberg, Scott
  Fleming, Michael~A Pfeffer, Jason Fries, and Nigam~H Shah.
\newblock The shaky foundations of clinical foundation models: A survey of
  large language models and foundation models for emrs.
\newblock {\em arXiv preprint arXiv:2303.12961}, 2023.

\bibitem{xie2022benchmarking}
Feng Xie, Jun Zhou, Jin~Wee Lee, Mingrui Tan, Siqi Li, Logasan~S/O Rajnthern,
  Marcel~Lucas Chee, Bibhas Chakraborty, An-Kwok~Ian Wong, Alon Dagan, et~al.
\newblock Benchmarking emergency department prediction models with machine
  learning and public electronic health records.
\newblock {\em Scientific Data}, 9(1):658, 2022.

\bibitem{yeche2021hirid}
Hugo Y{\`e}che, Rita Kuznetsova, Marc Zimmermann, Matthias H{\"u}ser, Xinrui
  Lyu, Martin Faltys, and Gunnar R{\"a}tsch.
\newblock Hirid-icu-benchmark--a comprehensive machine learning benchmark on
  high-resolution icu data.
\newblock {\em arXiv preprint arXiv:2111.08536}, 2021.

\end{thebibliography}
}

\newpage

\appendix

\section*{Supplementary Material}
\setcounter{page}{1}

%
%

\section{Author Responsibility Statement}
The authors confirm that they bear all responsibility in case of violation of rights or licenses.

\section{Public Accessibility + Licenses}
\label{section_licenses}

%
%

\subsection{Dataset}
\label{section_data_release}

We release {\sc EHRSHOT} under a research data use agreement. The dataset is available here: \href{https://ehrshot.stanford.edu/}{https://ehrshot.stanford.edu/}. Access is gated by a researcher data use agreement due to the sensitive nature of the dataset. We do not upload our dataset to another data repository due to these concerns. 

In order to ensure we do not reveal Protected Health Information (PHI) in our dataset, we take several precautions. First, we only release deidentified data. The deidentification process has been previously described in \cite{datta2020new}. Second, on top of this deidentification process, we also apply additional privacy-protecting transformations following the best practices of the MIMIC-III dataset \cite{johnson2016mimic}, which are detailed in Section \ref{section_data_preprocessing}. Third, we do not publish any clinical notes. Fourth, we release our dataset under a data usage agreement that requires researchers to register with their identity and gain approval before accessing the dataset.

\textbf{License:} The license for the dataset is the standard Stanford University Dataset Research Use Agreement, and is reproduced below:

\begin{displayquote}
    \tiny
    By registering for downloads from the EHRSHOT Dataset, you are agreeing to this Research Use Agreement, as well as to the Terms of Use of the Stanford University School of Medicine website as posted and updated periodically at http://www.stanford.edu/site/terms/. Permission is granted to view and use the EHRSHOT Dataset without charge for personal, non-commercial research purposes only. Any commercial use, sale, or other monetization is prohibited.
    
    Other than the rights granted herein, the Stanford University School of Medicine (“School of Medicine”) retains all rights, title, and interest in the EHRSHOT Dataset. You may make a verbatim copy of the EHRSHOT Dataset for personal, non-commercial research use as permitted in this Research Use Agreement. If another user within your organization wishes to use the EHRSHOT Dataset, they must register as an individual user and comply with all the terms of this Research Use Agreement. YOU MAY NOT DISTRIBUTE, PUBLISH, OR REPRODUCE A COPY of any portion or all of the EHRSHOT Dataset to others without specific prior written permission from the School of Medicine. YOU MAY NOT SHARE THE DOWNLOAD LINK to the EHRSHOT Dataset to others. If another user within your organization wishes to use the EHRSHOT Dataset, they must register as an individual user and comply with all the terms of this Research Use Agreement. You must not modify, reverse engineer, decompile, or create derivative works from the EHRSHOT Dataset. You must not remove or alter any copyright or other proprietary notices in the EHRSHOT Dataset.
    
    The EHRSHOT Dataset has not been reviewed or approved by the Food and Drug Administration, and is for non-clinical, Research Use Only. In no event shall data or images generated through the use of the EHRSHOT Dataset be used or relied upon in the diagnosis or provision of patient care. THE EHRSHOT Dataset IS PROVIDED "AS IS," AND STANFORD UNIVERSITY AND ITS COLLABORATORS DO NOT MAKE ANY WARRANTY, EXPRESS OR IMPLIED, INCLUDING BUT NOT LIMITED TO WARRANTIES OF MERCHANTABILITY AND FITNESS FOR A PARTICULAR PURPOSE, NOR DO THEY ASSUME ANY LIABILITY OR RESPONSIBILITY FOR THE USE OF THIS EHRSHOT Dataset. You will not make any attempt to re-identify any of the individual data subjects. Re-identification of individuals is strictly prohibited. Any re-identification of any individual data subject shall be immediately reported to the School of Medicine. Any violation of this Research Use Agreement or other impermissible use shall be grounds for immediate termination of use of this EHRSHOT Dataset. In the event that the School of Medicine determines that the recipient has violated this Research Use Agreement or other impermissible use has been made, the School of Medicine may direct that the undersigned data recipient immediately return all copies of the EHRSHOT Dataset and retain no copies thereof even if you did not cause the violation or impermissible use. In consideration for your agreement to the terms and conditions contained here, Stanford grants you permission to view and use the EHRSHOT Dataset for personal, non-commercial research. You may not otherwise copy, reproduce, retransmit, distribute, publish, commercially exploit or otherwise transfer any material.
    
    Limitation of Use: You may use EHRSHOT Dataset for legal purposes only. You agree to indemnify and hold Stanford harmless from any claims, losses or damages, including legal fees, arising out of or resulting from your use of the EHRSHOT Dataset or your violation or role in violation of these Terms. You agree to fully cooperate in Stanford’s defense against any such claims. These Terms shall be governed by and interpreted in accordance with the laws of California.
\end{displayquote}

%
%

\subsection{Pretrained Foundation Model (CLMBR-T-base)}
\label{section_model_release}

We release CLMBR-T-base, a foundation model pre-trained on the structured EHR data of roughly 2.5 million patients at Stanford Medicine \cite{steinberg2021language}. The model's weights can be found at our website here: \href{https://ehrshot.stanford.edu/}{https://ehrshot.stanford.edu/}. Access is gated by a researcher data use agreement due to the sensitive nature of the training dataset.

A concern with the release of such a model is the lack of solid theoretical privacy assurances, thus creating the possibility of the model revealing medical data. To mitigate these concerns, we implement several additional precautions. First, the model is trained exclusively on deidentified data to eliminate the chance of any Protected Health Information (PHI) seeping into the model. Second, all unique text strings released as part of our CLMBR-T-base model's dictionary (e.g. terms such as "Yes" or "No" that serve as categorical variables) were manually reviewed to ensure they do not reveal any PHI. Third, we make our model available under a data usage agreement that requires researchers to register with their identity and gain approval before accessing the model.

\textbf{License:} The license for the code for the model is here: \href{https://github.com/som-shahlab/femr/blob/main/LICENSE}{https://github.com/som-shahlab/femr/blob/main/LICENSE}. The license for the model weights is here: \href{https://huggingface.co/StanfordShahLab/clmbr-t-base}{https://huggingface.co/StanfordShahLab/clmbr-t-base}.

\section{Dataset Details}
\label{section_dataset_details}

\subsection{EHRSHOT Cohort}

Demographics of the EHRSHOT cohort are included below.

\begin{table}[h]
    \small
    \centering 
    \caption{{\sc EHRSHOT}: Patient demographics in the train, validation, and test splits.}
    \vspace{0.2cm}
    \begin{tabular}{llllll}
        \toprule
        \textbf{Attribute} & & \textbf{Train} & \textbf{Val} & \textbf{Test} & \textbf{All Splits}\\
        \midrule

        \multirow{2}{*}{\textbf{Gender}} 
        & Male & 1122 & 1090 & 1086 & 3298 \\
        & Female & 1173 & 1142 & 1126 & 3441 \\

        \multirow{5}{*}{\textbf{Age}} 
        & 19-20 & 8 & 3 & 2 & 13 \\
        & 21-40 & 412 & 457 & 431 & 1300 \\
        & 41-60 & 648 & 597 & 576 & 1821  \\
        & 61-80 & 916 & 892 & 905 & 2713 \\
        & 81-88 & 311 & 283 & 298 & 892 \\~\\
        
        \multirow{6}{*}{\textbf{Race}} 
        & American Indian & 14 & 7 & 4 & 25 \\
        & Asian & 356 & 347 & 340 & 1043 \\
        & Black & 98 & 105 & 95 & 298 \\
        & Pacific Islander & 23 & 21 & 30 & 74 \\
        & White & 1286 & 1222 & 1228 & 3736 \\
        & Unknown & 518 & 530 & 515 & 1563 \\~\\
        
        \multirow{2}{*}{\textbf{Ethnicity}} 
        & Hispanic & 374 & 342 & 322 & 1038 \\
        & Non-Hispanic & 1921 & 1890 & 1890 & 5701 \\~\\

        \textbf{Total} &  & 2295 & 2232 & 2212 & 6739 \\
        \bottomrule
    \end{tabular}
    \label{tab:demographics} 
\end{table}

\begin{figure}[h]
    \centering
    \caption{Histograms of {\sc EHRSHOT} patient timeline characteristics}
    \begin{subfigure}[b]{\textwidth}
        \centering
        \includegraphics[width=1\textwidth]{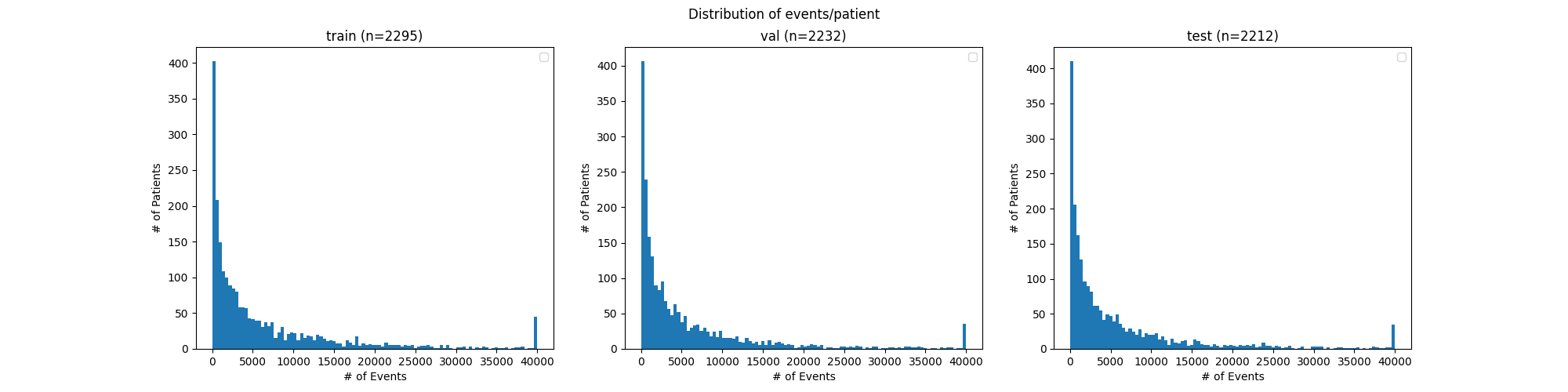}
        \caption{Total number of events per patient, broken down by train/val/test split. Note that the x-axis is clamped at 40000 for clarity (i.e. along the x-axis we plot $\text{min}(x, 40000)$)}.
        \label{fig:events_per_patient}
    \end{subfigure}
    \begin{subfigure}[b]{\textwidth}
        \centering
        \includegraphics[width=1\textwidth]{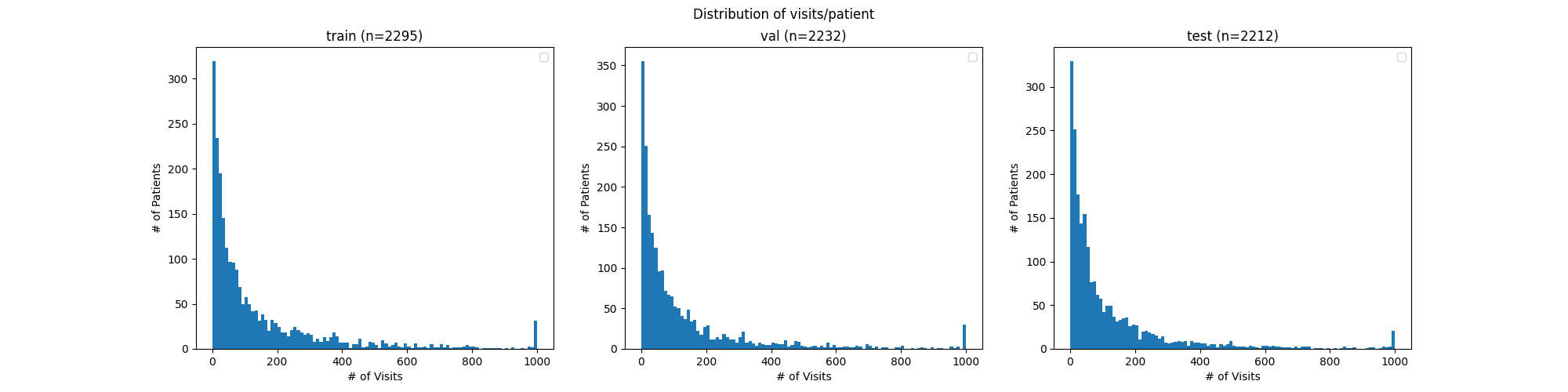}
        \caption{Total number of visits per patient, broken down by train/val/test split. Note that the x-axis is clamped at 1000 for clarity (i.e. along the x-axis we plot $\text{min}(x, 1000)$)}.
        \label{fig:visits_per_patient}
    \end{subfigure}
    \begin{subfigure}[b]{\textwidth}
        \centering
        \includegraphics[width=1\textwidth]{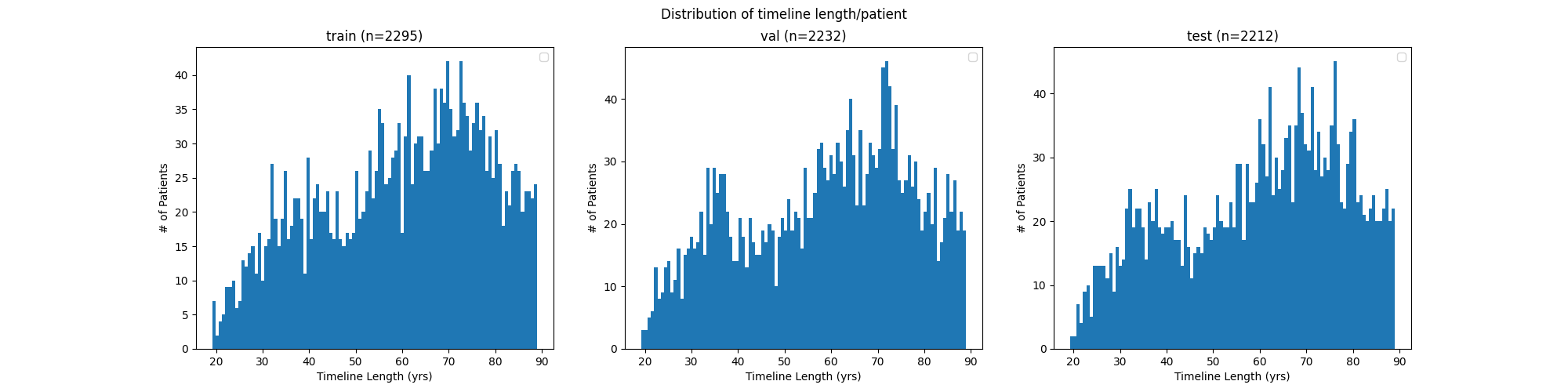}
        \caption{Total length of each patient timeline (i.e. difference in time between birth date and last recorded event), broken down by train/val/test split.}
        \label{fig:timeline_length_per_patient_years}
    \end{subfigure}
    \label{fig:timeline_histograms}
\end{figure}

\subsection{Pretraining Dataset}
The pretraining dataset for CLMBR-T-base contains a total of 3.67 million patient records, of which 2.57 million are used to train the model. We include summary statistics of these patients' demographics in Table \ref{tab:pretrain_demographics} and Table \ref{tab:pretrain_patient_summary_statistics}.

\begin{table}[h]
    \small
    \centering 
    \caption{Pretraining Dataset: Patient demographics in the train, validation, and test splits.}
    \vspace{0.2cm}
    \begin{tabular}{llllll}
        \toprule
        \textbf{Attribute} & & \textbf{Train} & \textbf{Val} & \textbf{Test} & \textbf{All Splits}\\
        \midrule

        \multirow{2}{*}{\textbf{Gender}} 
        & Male & 1186614 & 255179 & 254733 & 1696526 \\
        & Female & 1380836 & 295126 & 296127 & 1972089 \\~\\

        \multirow{5}{*}{\textbf{Age}} 
        & 0-20 & 625949 & 135045 & 134684 & 895678 \\
        & 21-40 & 671018 & 143607 & 144045 & 958670 \\
        & 41-60 & 617519 & 131996 & 132432 & 881947 \\
        & 61-80 & 502842 & 107299 & 107746 & 717887 \\
        & 81-88 & 150122 & 32358 & 31953 & 214433 \\~\\
        
        \multirow{6}{*}{\textbf{Race}} 
        & American Indian & 7229 & 1509 & 1516 & 10254 \\
        & Asian & 371065 & 79638 & 80418 & 531121 \\
        & Black & 83624 & 17895 & 17919 & 119438 \\
        & Pacific Islander & 20959 & 4350 & 4435 & 29744 \\
        & White & 987676 & 211262 & 211429 & 1410367 \\
        & Unknown & 1096897 & 235651 & 235143 & 1567691 \\~\\
        
        \multirow{2}{*}{\textbf{Ethnicity}} 
        & Hispanic & 325037 & 69912 & 69689 & 464638 \\
        & Non-Hispanic & 2242413 & 480393 & 481171 & 3203977 \\~\\

        \textbf{Total} &  & 2567450 & 550305 & 550860 & 3668615 \\
        \bottomrule
    \end{tabular}
    \label{tab:pretrain_demographics} 
\end{table}

\begin{table}[h]
    \small
    \centering 
    \caption{Pretraining Dataset: Summary statistics on the number of events, visits, and length of patient timelines.}
    \vspace{0.2cm}
        
    \begin{tabular}{llllll}
        \toprule
        \textbf{Attribute} & & \textbf{Train} & \textbf{Val} & \textbf{Test} & \textbf{All Splits}\\
        \midrule
        \multirow{3}{*}{\textbf{Number of Events}} 
            & Min  & 1 & 1 & 1 & 1 \\
            & Mean & 707 & 706 & 704 & 706 \\
            & Max  & 191369 & 213133 & 214400 & 214400 \\~\\
        \multirow{3}{*}{\textbf{Number of Visits}} 
            & Min  & 0 & 0 & 0 & 0 \\
            & Mean & 28 & 28 & 28 & 28\\
            & Max  & 3701 & 4305 & 3109 & 4305 \\~\\
        \multirow{3}{*}{\textbf{Timeline Length (yrs)}} 
            & Min  & 0 & 0 & 0 & 0 \\
            & Mean & 40 & 40 & 40 & 40 \\
            & Max  & 92 & 90 & 90 & 92 \\
        \bottomrule
    \end{tabular}
    \label{tab:pretrain_patient_summary_statistics} 
\end{table}

\subsection{Task Definitions}
\label{section_task_definitions}

Here, we detail the precise definitions for each of the 15 tasks for which we provide labels in our benchmark dataset.

\textbf{Operational Outcomes}. These tasks are related to hospital operations. They are all binary classification tasks, and are defined as follows: 

\begin{itemize}
    \item \textbf{Long Length of Stay}: Predict whether a patient's total length of stay during a visit to the hospital will be at least 7 days. The prediction time is at 11:59pm on the day of admission, and visits that last less than one day (i.e. discharge occurs on the same day of admission) are ignored.
    \item \textbf{30-day Readmission}: Predict whether a patient will be re-admitted to the hospital within 30 days after being discharged from a visit. The prediction time is at 11:59pm on the day of admission, and admissions where a readmission occurs on the same day as the corresponding discharge are ignored.
    \item \textbf{ICU Transfer}: Predict whether a patient will be transferred to the ICU during a visit to the hospital. The prediction time is at 11:59pm on the day of admission, and ICU transfers that occur on the same day as admission are ignored.
\end{itemize}

\textbf{Anticipating Lab Test Results}. These tasks are related to lab value prediction. They are all multiclass classification tasks. The prediction time is immediately before the lab result is recorded. They are defined as follows: 

\begin{itemize}
    \item \textbf{Thrombocytopenia}: Predict whether a thrombocytopenia lab comes back as normal (>=150 $10^9$/L), mild (>=100 and <150 $10^9$/L), moderate (>=50 and <100 $10^9$/L), or severe (<50 $10^9$/L),. We consider all lab results coded as LOINC/LP393218-5, LOINC/LG32892-8, or LOINC/777-3.
    \item \textbf{Hyperkalemia}: Predict whether a hyperkalemia lab comes back as normal (<=5.5 mmol/L), mild (>5.5 and <=6mmol/L), moderate (>6 and <=7 mmol/L), or severe (>7 mmol/L). We consider all lab results coded as LOINC/LG7931-1, LOINC/LP386618-5, LOINC/LG10990-6, LOINC/6298-4, or LOINC/2823-3.
    \item \textbf{Hypoglycemia}: Predict whether a hypoglycemia lab comes back as normal (>=3.9 mmol/L), mild (>=3.5 and <3.9 mmol/L), moderate (>=3 and <3.5 mmol/L), or severe (<3 mmol/L). We consider all lab results coded as SNOMED/33747003, LOINC/LP416145-3, or LOINC/14749-6.
    \item \textbf{Hyponatremia}: Predict whether a hyponatremia lab comes back as normal (>=135 mmol/L), mild (>=130 and <135 mmol/L), moderate (>=125 and <130 mmol/L), or severe (<125 mmol/L). We consider all lab results coded as LOINC/LG11363-5, LOINC/2951-2, or LOINC/2947-0.
    \item \textbf{Anemia}: Predict whether an anemia lab comes back as normal (>=120 g/L), mild (>=110 and <120 g/L), moderate (>=70 and <110 g/L), or severe (<70 g/L). We consider all lab results coded as LOINC/LP392452-1.
\end{itemize}

Please note that for the results of our baseline experiments in Section \ref{section_results}, we reframe these lab value tasks as binary classification tasks, where a label is "negative" if the result is normal and "positive" otherwise.

\begin{table}[h]
  \small
  \centering 
  \caption{Task Prediction Windows. \textit{Prediction Time} is the precise time point (up to minute precision) in a patient's timeline when the prediction is made. \textit{Time Horizon} is the length of time considered after the prediction time  to determine whether an event occurs, i.e. we only consider a patient "positive" for a new diagnosis of pancreatic cancer if she receives that diagnosis within a year of being discharged.}
  \vspace{0.2cm}  
  \begin{tabularx}{\textwidth}{p{4cm} p{1.5cm} p{4cm} p{3cm}}
  \toprule
    \textbf{Task Name} & \textbf{Task Type} & \textbf{Prediction Time} & \textbf{Time Horizon} \\
    \midrule
    \multicolumn{2}{l}{\textbf{Operational Outcomes}}\\
    Long Length of Stay & Binary & 11:59pm on day of admission & Admission duration \\ 
    30-day Readmission & Binary & 11:59pm on day of discharge & 30 days post-discharge \\ 
    ICU Transfer & Binary & 11:59pm on day of admission & Admission duration \\~\\
    \multicolumn{2}{l}{\textbf{Anticipating Lab Test Results}}\\
    Thrombocytopenia & 4-way \newline multiclass & Immediately before result & Next result \\ 
    Hyperkalemia & 4-way \newline multiclass & Immediately before result & Next result \\ 
    Hypoglycemia & 4-way \newline multiclass & Immediately before result & Next result \\ 
    Hyponatremia & 4-way \newline multiclass & Immediately before result & Next result \\ 
    Anemia & 4-way \newline multiclass & Immediately before result & Next result \\~\\
    \multicolumn{2}{l}{\textbf{Assignment of New Diagnoses}}\\
    Hypertension & Binary & 11:59pm on day of discharge & 1 year post-discharge \\ 
    Hyperlipidemia & Binary & 11:59pm on day of discharge & 1 year post-discharge \\ 
    Pancreatic Cancer & Binary & 11:59pm on day of discharge & 1 year post-discharge \\ 
    Celiac & Binary & 11:59pm on day of discharge & 1 year post-discharge \\ 
    Lupus & Binary & 11:59pm on day of discharge & 1 year post-discharge \\ 
    Acute MI & Binary & 11:59pm on day of discharge & 1 year post-discharge \\~\\ 
    \multicolumn{2}{l}{\textbf{Anticipating Chest X-ray Findings}}\\
    Chest X-Ray Findings & 14-way \newline multilabel & 24hrs before report is recorded & Next report \\ 
    \bottomrule
  \end{tabularx}
  \label{tab:task_definitions} 
\end{table}

\textbf{Assignment of New Diagnoses}. These tasks are related to predicting the first diagnosis of a disease. They are all binary classification tasks. The prediction time is at 11:59pm on the day of discharge from an inpatient visit, and we count any diagnosis that occurs within 365 days post-discharge as a positive outcome. We ignore all discharges in which the patient already has an existing diagnosis of a disease. The tasks are defined as follows: 

\begin{itemize}
    \item \textbf{Hypertension}: Predict whether the patient will have her first diagnosis of essential hypertension within the next year. We define hypertension as an occurrence of the code SNOMED/59621000, as well as its children codes in our ontology.
    \item \textbf{Hyperlipidemia}: Predict whether the patient will have her first diagnosis of hyperlipidemia within the next year. We define hyperlipidemia as an occurrence of the code SNOMED/55822004, as well as its children codes in our ontology.
    \item \textbf{Pancreatic Cancer}: Predict whether the patient will have her first diagnosis of pancreatic cancer within the next year. We define pancreatic cancer as an occurrence of the code SNOMED/372003004, as well as its children codes in our ontology.
    \item \textbf{Celiac}: Predict whether the patient will have her first diagnosis of celiac disease within the next year. We define celiac disease as an occurrence of the code SNOMED/396331005, as well as its children codes in our ontology.
    \item \textbf{Lupus}: Predict whether the patient will have her first diagnosis of lupus within the next year. We define lupus as an occurrence of the code SNOMED/55464009, as well as its children codes in our ontology.
    \item \textbf{Acute MI}: Predict whether the patient will have her first diagnosis of an acute myocardial infarction within the next year. We define myocardial infarction as an occurrence of the code SNOMED/57054005, as well as its children codes in our ontology.
\end{itemize}

\textbf{Anticipating Chest X-ray Findings}. The chest X-ray findings task is a multilabel classification task to identify which of 14 possible findings were included in a chest X-ray report. The prediction time is 24 hours before the radiology report is recorded. The labels are derived by running the CheXpert NLP labeler on the unstructured text of the corresponding radiology report \cite{irvin2019chexpert}. We do not release this unstructured text as part of our dataset due to patient privacy concerns.

The possible findings are as follows: "No Finding", "Enlarged Cardiomediastinum", "Cardiomegaly", "Lung Lesion", "Lung Opacity", "Edema", "Consolidation", "Pneumonia", "Atelectasis", "Pneumothorax", "Pleural Effusion", "Pleural Other", "Fracture", "Support Devices".

\subsection{Dataset Format}
\label{section_data_format}

Our dataset is comprised of two main sets of tabular files: \textbf{(A) Events} files which contain all of the clinical events associated with every patient in our dataset, and \textbf{(B) Labels} files which contain the labels associated with all of our benchmark tasks for every patient in our dataset.

\textbf{(A) Events} is as a set of CSV files containing every clinical event that happened to the patients in our dataset. Every row is a unique clinical event. Each CSV file shares the same column schema, which is as follows:

\begin{itemize}
    \item \textbf{Patient ID} - Integer - Unique identifier for patient
    \item \textbf{Start} - Datetime - Start time of event
    \item \textbf{End} - Datetime (optional) - End time of event
    \item \textbf{Code} - String - Name of the clinical event (e.g. "SNOMED/3950001" or "ICD10/I25.110")
    \item \textbf{Value} - Float/String (optional) - Either a numerical value associated with an event (e.g. a lab test result) or a string associated with a categorical variable (e.g. "Yes/No" questions)
    \item \textbf{Unit} - String (optional) - Unit of measurement for \textbf{Value}
    \item \textbf{Visit ID} - Integer (optional) - Unique identifier for the visit during which this event occurred
    \item \textbf{OMOP-CDM Table} - String - Name of the source OMOP-CDM table where this event was recorded
\end{itemize}

Every event is associated with the OMOP-CDM table in the source STARR database from which it was taken (\textbf{OMOP-CDM Table}) \cite{omop-cdm}. Researchers unfamiliar with the OMOP-CDM can simply ignore this column.

\textbf{(B) Labels} is a set of CSV files containing the labels for every task for every patient. Every row is a unique label associated with a specific patient, task, and time point. Each CSV file shares the same column schema, which is as follows:

\begin{itemize}
    \item \textbf{Patient ID} - Integer - Unique identifier for patient
    \item \textbf{Prediction Time} - Datetime - Time at which the prediction for this label is made
    \item \textbf{Value} - Boolean / Integer - Value for this label. Boolean if task is binary classification. Integer if task is multiclass or multilabel classification.
    \item \textbf{Label Type} - String - Type of task associated with this label. Can be "boolean" (binary classification), "numeric" (regression), "survival" (time-to-event), or "categorical" (multilabel or multiclass classification).
\end{itemize}

\subsection{Data Preprocessing}
\label{section_data_preprocessing}

The source dataset we use, the Stanford STARR research database \cite{datta2020new}, is an Observational Medical Outcomes Partnership Common Data Model (OMOP-CDM) \cite{hripcsak2015observational} compliant transformation of data extracted from Stanford's production EHR system (Epic). We do not alter any of the transformations or deidentifiation steps in the ETL used to generate this OMOP-CDM extract described in \cite{datta2020new}. 

Following the best practices of the MIMIC-III dataset, we apply several additional custom transformations to prevent data leakage and add an additional layer of patient privacy protections \cite{johnson2016mimic}. First, we jitter all dates within each patient timeline by the same random amount (to a random year between 2100 and 2200). Second, we remove all patients <= 18 or >= 89 years of age. Third, we remove all instances of free form text (i.e., notes and narratives). For the clinical events which take on categorical values specified as strings (e.g. a questionnaire which can be answered "Yes" or "No"), we select the top-100 most representative such categorical text strings, manually verify that they do not contain any PHI, and remove the rest of the text strings from our model release by replacing them with blank strings. This preserves roughly 65\% of all categorical values in our dataset. Fourth, we remove any patients with less than 10 clinical events in their record. Fifth, we adjust the timing of certain events to more realistically reflect the chronology of care delivery. Specifically, we move any events recorded before a patient's birth to after their time of birth; we set the start times of visits equal to the start time of the first event in each visit; we move billing codes recorded during a visit to the end of the visit; we move any event coded at midnight to 11:59pm of that day; we remove all duplicate codes that occur sequentially on the same day; and we remove all codes with `None` values that occur on the same day as an identical code with a non-`None` value associated with it. These transformations are all specified in code \href{https://github.com/som-shahlab/ehrshot-benchmark/}{in our Github repo}.

\subsection{Cohort Selection Process}
\label{section_cohort_generation}

We selected a cohort of 6,739 patients for {\sc EHRSHOT} from the larger STARR source dataset of 3.67M patients. Per the motivation of this project, we were primarily interested in few-shot evaluation of models across diverse tasks. Several of the tasks that would be of interest to a health system, however, have fairly low prevalence within the general patient population. Thus, we needed to construct our cohort in a way that preserved sufficient positive labels to enable downstream models to conduct few-shot learning. We aimed to have at least $k = 128$ positive and negative examples in each of the train/val/test splits for every task that we considered in order to allow for a broad range of few-shot learning scenarios, and at least $k = 128$ positive examples for each label within a multiclass or multilabel classification task. Where this was not possible (e.g. the Celiac task), we included as many positive labels in each split as possible.

We began with our set of 15 tasks of interest. For each task, we labeled all patients within our source database per that task's definition. For tasks that have a low prevalence (which we consider as a 1:5 ratio of positive to negative labels), we subsample negative labels to bring the prevalence of positive labels up to that ratio. We then subsample further for few-shot evaluation, selecting 128 unique patients for each split who have at least one positive label for the task. We then sample sufficient negative labels to maintain the chosen prevalence. We repeat this process for all tasks to arrive at our final cohort of patients. For each successive task, we prioritize selecting patients that have already been sampled into our cohort to reduce the total number of patients added to our cohort (since some patients have positive labels for multiple tasks).

\section{Results Details}
\label{section_results_details}

In order to fully reproduce our results, please follow the instructions at our Github repo here: \href{https://github.com/som-shahlab/ehrshot-benchmark}{https://github.com/som-shahlab/ehrshot-benchmark}.

\subsection{Problem Formulation}

Our dataset and models can be formulated as follows. Our dataset $\mathcal{D} = (\{ (\mathbf{X}_p, \mathbf{Y}_p) \}_{p = 1}^{|\mathcal{P}|}$ contains the full coded medical timeline ($\mathbf{X}_p$) and task-specific set of labels ($\mathbf{Y}_p$) for each patient $p \in \mathcal{P}$, for a total of $|\mathcal{P}|$ patients. Each patient $p$ is defined by a sequence of clinical events $\mathbf{X}_p = \{ x_{p1}, x_{p2}, ..., x_{pn} \}$, where $x_{pi}$ denotes the $i$th code in the timeline of patient $p$. Note that a code $x_{pi}$ can be any form of structured data taken from the patient's EHR, including a diagnosis, procedure, medication prescription, lab test, etc. We define $\mathbf{X}_p^{(t)}$ to be the patient timeline up to time $t$ -- i.e. if event $x_{pj}$ occurs before or at $t$ but $x_{p(j+1)}$ occurs after $t$, then if $\mathbf{X}_p = \{ x_{p1}, ..., x_{pj}, x_{p(j+1)}, ..., x_{pn} \}$ we have that $\mathbf{X}_p^{(t)} = \{ x_{p1},..., x_{pj} \}$. 

In addition to the timeline of each patient, our dataset also contains labels for each task and patient. We define benchmark tasks $b \in \mathcal{B}$, where $|\mathcal{B}| = 15$ for our dataset. Each patient has a set of labels $\mathbf{Y}_p = \{ y^{(t_1)}_{p b_1}, y^{(t_2)}_{p b_1},..., y^{(t_L)}_{p b_{|\mathcal{B}|}} \}$, where $L$ is the total number of labels for patient $p$, and  the expression $y^{(t_j)}_{p b_i}$ represents the label for patient $p$ for task $b_i$ at time point $t_j$.

We are interested in making predictions of the following format: Given a patient $p$'s entire medical history up to and including time point $t$ (i.e. $\mathbf{X}_p^{(t)}$), predict the value of $y^{(t)}_{p b}$ for each corresponding benchmark task $b \in \mathcal{B}$ where such a label exists.

Please note that this prediction task is at the level of individual clinical events rather than visits/encounters.

\subsection{Count-Based GBM}

We train a LightGBM model as one of our baseline models. In order to train such a model on a patient's timeline, we must first featurize the timeline into a vector. We follow best practices for competitive baseline models by using count-based featurization, in which a patient is transformed into a vector containing the counts of how many times each clinical event has occurred in that patient's timeline prior to the prediction time point \cite{rajkomar2018scalable, steinberg2021language}. 

Let $\mathcal{C}$ be the set of all unique medical codes in our dataset. Let us consider making a prediction for patient $p$ at time $t$. Then the count-based featurization for $p$ at time $t$ is given by the vector $\mathbf{p}^{(t)} \in \mathbb{N}^{|\mathcal{C}|}$, where each element is defined as $\mathbf{p}^{(t)}_i = \sum_{x_j \in \mathbf{X}^{(t)}_p} I(x_j = i)$, i.e. the count of medical code $i$ recorded for patient $p$ before the prediction time $t$. Stacking these patient vectors results in a count matrix $\mathbf{M} \in \mathbb{N}^{|\mathbf{Y}| \times |\mathcal{C}|}$. As there are hundreds of thousands of unique codes, most of which occur infrequently among patients, this results in a very high-dimensional and sparse matrix. 

To help address the sparseness of $\mathbf{M}$, we use a technique called \textit{ontology expansion} \cite{choi2017gram}, in which we count each occurrence of a code once for the code itself, and once for every parent node of that code in the OMOP ontology up to the root node of our ontology. Consider the ICD10 code E10.1 (Type 1 diabetes mellitus). Any occurrence of this code in a patient's timeline should also give the patient "credit" for having the parent codes of E10.1 -- E10 (Type 1 diabetes mellitus) and E08–E13 (Diabetes mellitus). This is because having E10.1 implies that the patient has E10 and E08-E13. We leverage existing OMOP ontology tools for ontology expansion and map codes to their ancestors. Then, when constructing our count matrix $\mathbf{M}$, we count each occurrence of a code for both that code and all of its parent codes. We refer to this ontology-expanded version of our count matrix as $\mathbf{M'}$.

Once the ontology-expanded count matrix $\mathbf{M'}$ is generated, a LightGBM model is trained on this input to predict the target label for each task \cite{ke2017lightgbm}. Hyperparameter tuning is performed on a validation set following the schedule described in Table \ref{tab:gbm_hyperparameters}.

\subsection{CLMBR-T-base}

For CLMBR-T-base, each unique medical code $c \in \mathcal{C}$ is associated with a $d$-dimensional embedding $e^c \in \mathbb{R}^d$. Each medical code $x_{pi} = c$ in patient $p$'s timeline is associated with both a code embedding $e^c$ and a position embedding $e^s$ which is defined using rotary positions embeddings \cite{su2021roformer}. Thus, the input to the model for $x_{pi}$ is given by the concatenation of these vectors, i.e. $e^c \mathbin\Vert e^s$. For our model, the code embeddings $e^c$ are generated using a standard embedding layer with a vocabulary size of $|\mathcal{C}| = 65,536$. Though there are more unique codes in our dataset, we only keep the top 65,536 codes with the highest contribution to the overall entropy of the dataset -- the rest of the codes occurring in our dataset are discarded in order to keep the size of our model's dictionary tractable. 

Lab values were discretized by computing decile statistics over the entire dataset and then creating tokens for each lab / decile pair. For example, if the 40th percentile of weight is 180 pounds and the 50th percentile is 190 pounds, we would create one token for “Weight/180-190” which would represent all events with values in that range.

Given this fixed dictionary, a classification task is defined to predict the next code in a patient's timeline given their preceding codes. We use a transformer as our classification model. Our transformer uses a local attention mechanism with a fixed context window of 496 tokens (i.e. clinical events) per layer. As our CLMBR model contains 12 stacked layers, this gives our model an effective context window of 496 * 12 = 5,952 clinical events, on which it conditions to generate its output at each step $i$. Sequences longer than that were truncated.

The output at step $i$ is a $d$-dimensional vector representation of the cumulative information up to and including event $x_{pi}$. We stack these representations for patient $p$ into a matrix $\mathbf{R_p} \in \mathbb{R}^{|\mathbf{X_p}| \times d}$ such that $\mathbf{R_p}_i$ is the cumulative $d$-dimensional representation of all events up to and including event $i$ for patient $p$. We then take the dot product of each row in this matrix with every code embedding $e^j$ for all $j \in \mathcal{C}$ in order to calculate a logit for each code $j$ at each event $i$, thus yielding: $\text{logit}_{pij} = \mathbf{R_p}_i \cdot e^j$. The model is then trained end-to-end using standard cross-entropy classification log-likelihood loss, employing an indicator variable $I_{pij}$ to mark if the next event for patient $p$ after event $i$ is an event with code $j$.

The overall loss function, $L(I | \text{logit})$, is computed as:

$$
    L(I | \text{logit}) = \prod_{p, i, j} I_{pij} \cdot  \text{softmax}(\text{logit}_{pi})_j
$$

For training our model, we use the best hyperparameters identified in \cite{guo2023ehr} and perform a limited hyperparameter search as defined in \ref{tab:clmbr_hyperparameters}.

\begin{figure}[htbp]
    \centering
    \includegraphics[width=\textwidth]{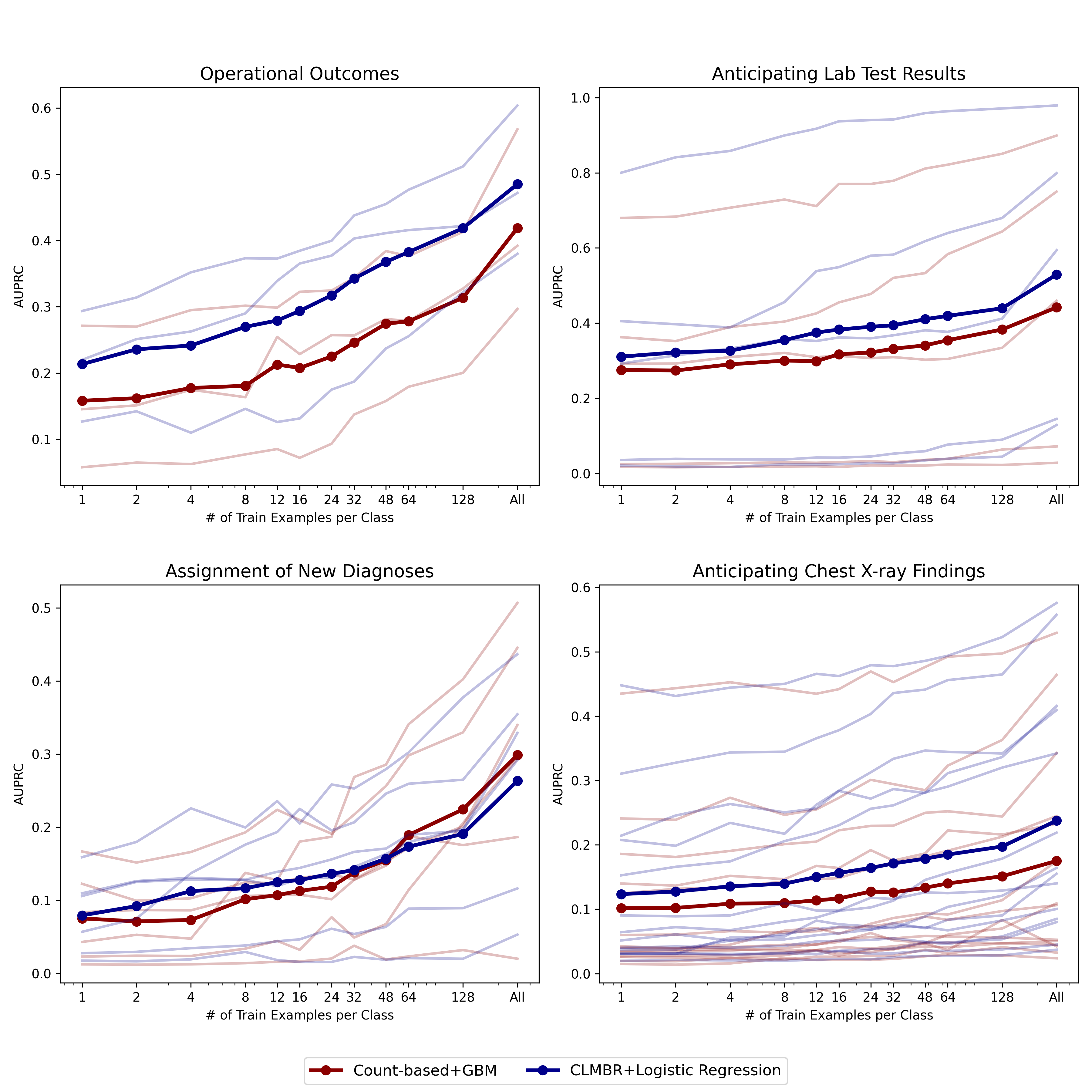}
    \caption{Aggregated AUPRC across all subtasks within each of the 4 task categories for $k \in \{1, 2, 4, 8, 12, 16, 24, 32, 48, 64, 128 \}$ shots. We also show performance on the full training set as \textit{All}. The \textbf{bolded lines} are the Macro-AUPRC for each model, averaged across all subtasks within a task category for each value of $k$. The blurred lines are the average AUPRC across 5 replicates for each subtask within a task category. Similar to the case with AUROC, the pretrained foundation model CLMBR-T-base (\textcolor{blue}{blue}) performs better across all $k$ on the \textit{Operational Outcomes}, \textit{Anticipating Lab Test Results}, and \textit{Anticipating Chest X-ray Findings} tasks, while the count-based GBM model (\textcolor{red}{red}) performs slightly better at higher $k$ on the \textit{Assignment of New Diagnoses} tasks.}
    \label{fig:agg_auprc}
\end{figure}

\begin{figure}[htbp]
    \centering
    \includegraphics[width=\textwidth]{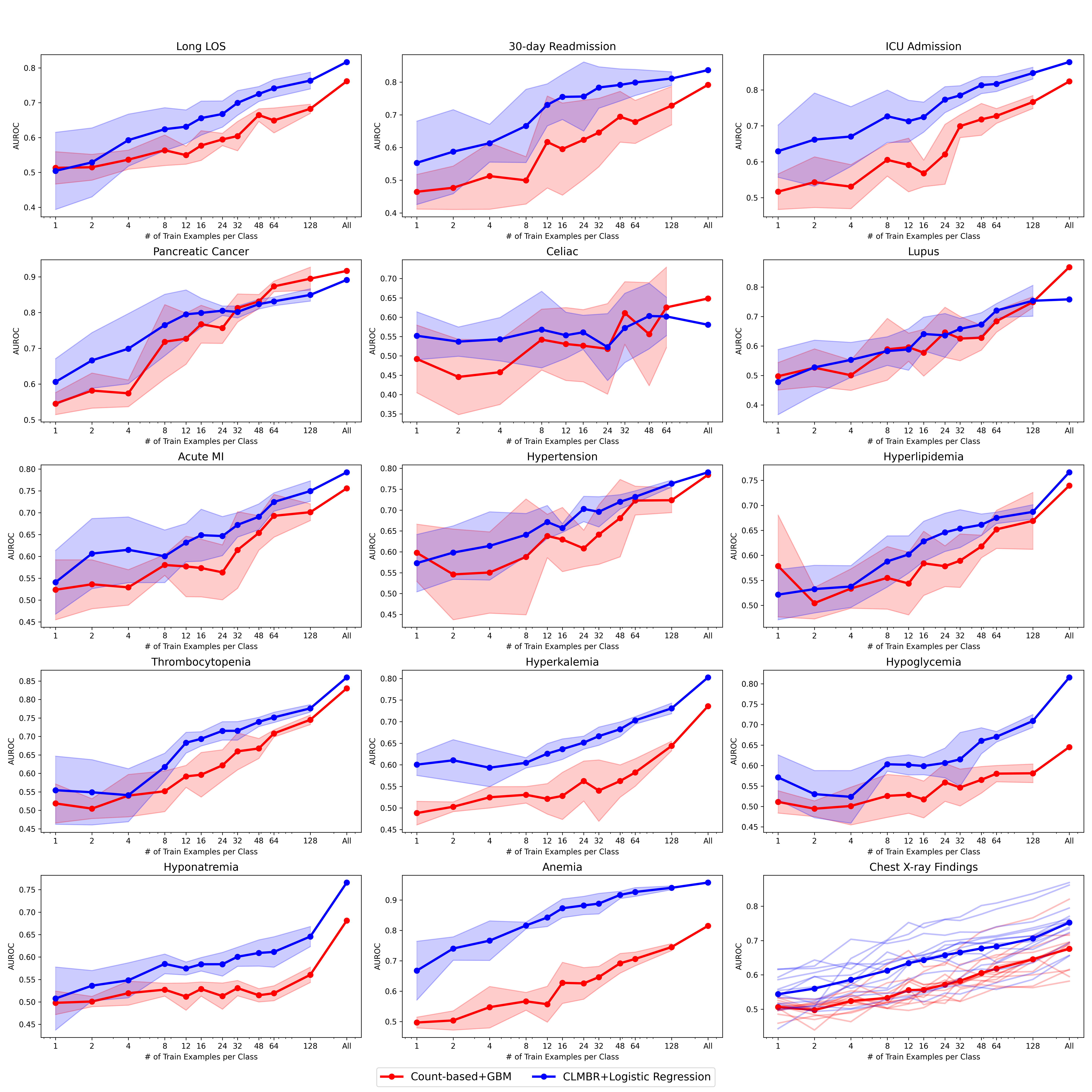}
    \caption{AUROC scores for each model across $k \in \{1, 2, 4, 8, 12, 16, 24, 32, 48, 64, 128\}$ shots. We also show performance on the full training set as \textit{All}. The pretrained foundation model CLMBR-T-base (\textcolor{blue}{blue}) shows stronger performance on \textit{Operational Outcomes} and \textit{Anticipating Lab Test Results} tasks, while the count-based GBM model (\textcolor{red}{red}) exhibits competitive performance at higher values of $k$ for the \textit{Assignment of New Diagnoses} tasks. For \textit{Chest X-ray Findings}, each blurred line represents one of the 14 individual labels, and the bolded line is macro-AUROC across all labels.}
    \label{fig:separate_auroc}
\end{figure}

\begin{figure}[htbp]
    \centering
    \includegraphics[width=\textwidth]{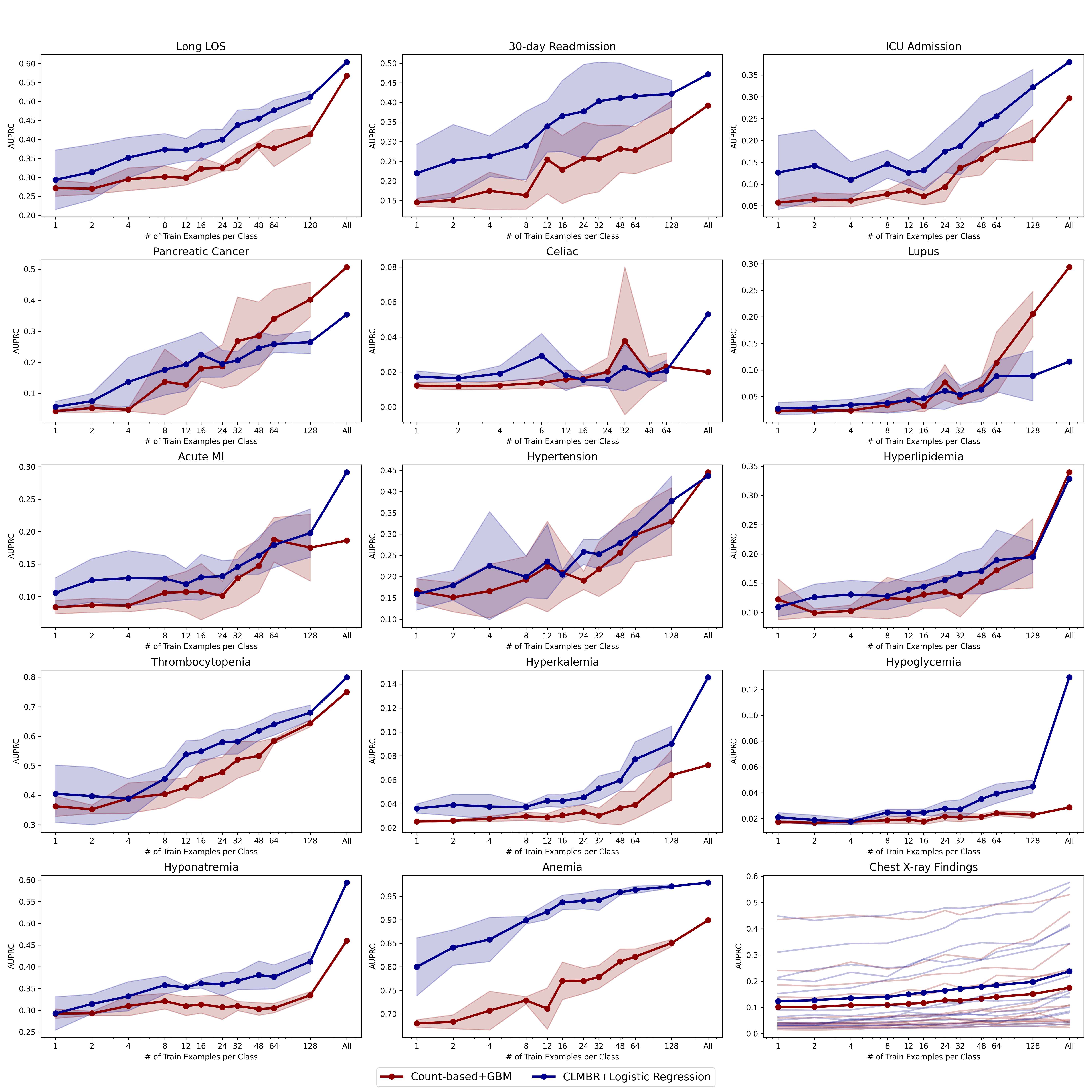}
    \caption{AUPRC scores for each model across $k \in \{1, 2, 4, 8, 12, 16, 24, 32, 48, 64, 128\}$ shots. We also show performance on the full training set as \textit{All}. CLMBR-T-base is in \textcolor{blue}{blue}, count-based GBM model is in \textcolor{red}{red}. For \textit{Chest X-ray Findings}, each blurred line represents one of the 14 individual labels, and the bold line is macro-AUPRC across all labels.}
    \label{fig:separate_auprc}
\end{figure}

\begin{figure}[htbp]
    \centering
    \includegraphics[width=\textwidth]{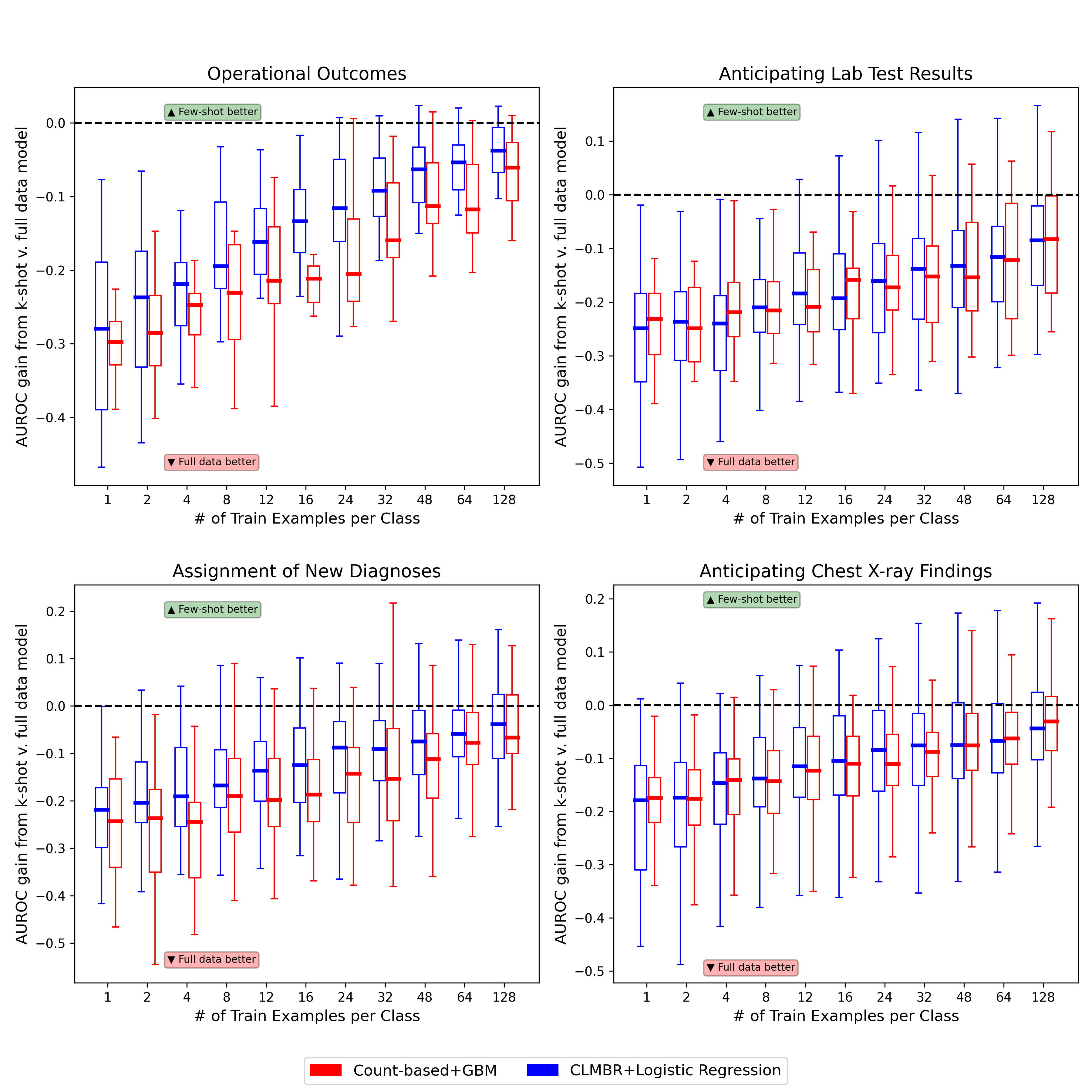}
    \caption{Difference in AUROC between each $k$-shot model replicate and a model trained on the full dataset. The pretrained foundation model CLMBR-T-base (\textcolor{blue}{blue}) closes the gap with the full data model faster than does the count-based GBM model (\textcolor{red}{red}).}
    \label{fig:full_comparison_auroc}
\end{figure}

\begin{figure}[htbp]
    \includegraphics[width=\textwidth]{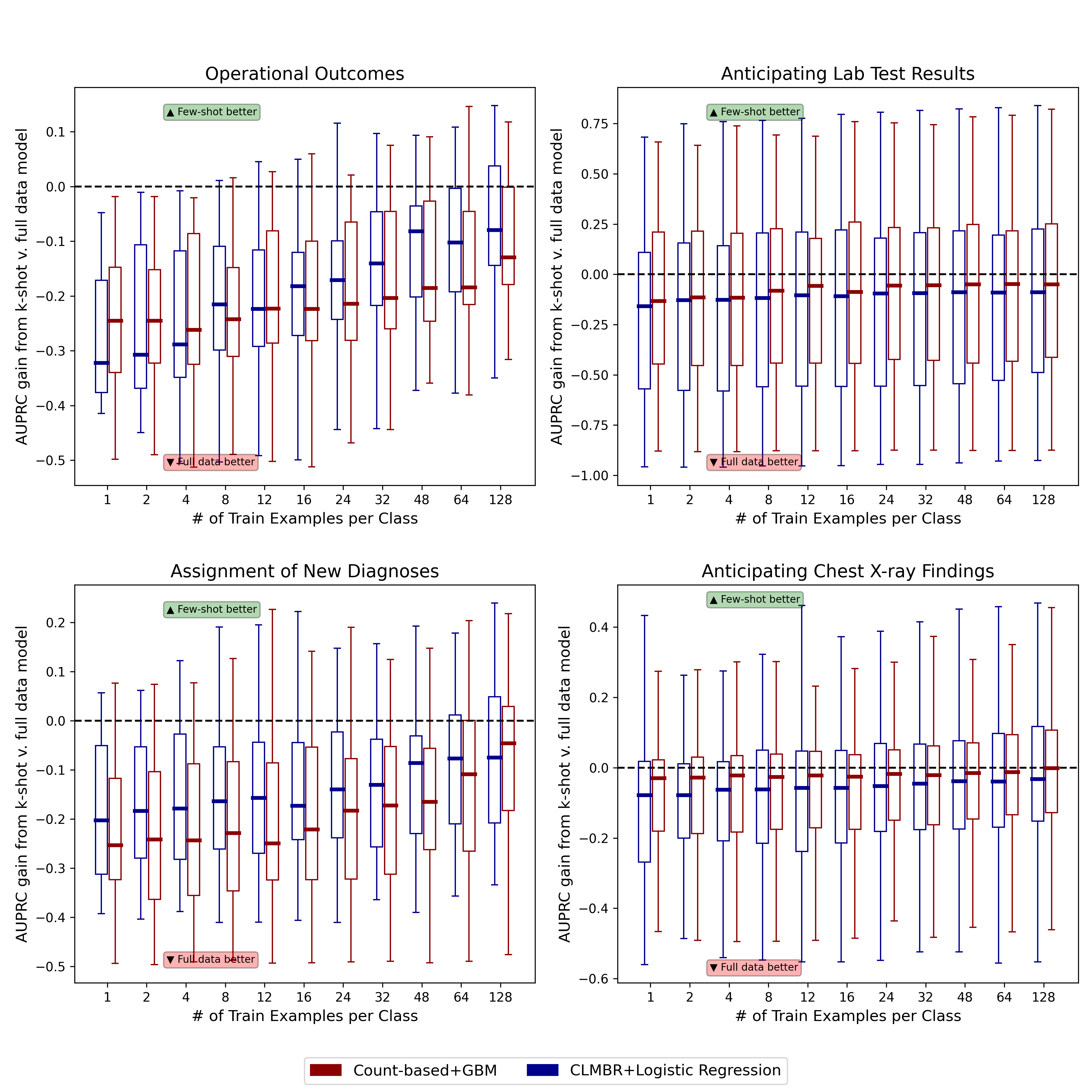}
    \caption{Difference in AURPC between each $k$-shot model replicate and a model trained on the full dataset. The pretrained foundation model CLMBR-T-base (\textcolor{blue}{blue}) closes the gap with the full data model faster than does the count-based GBM model (\textcolor{red}{red}).}
    \label{fig:full_comparison_auprc}
\end{figure}

\definecolor{darkyellow}{rgb}{1.0, 0.75, 0.0}
\definecolor{darkgreen}{rgb}{0.0, 0.5, 0.0}

\begin{figure}[htbp]
    \centering
    \includegraphics[width=\textwidth]{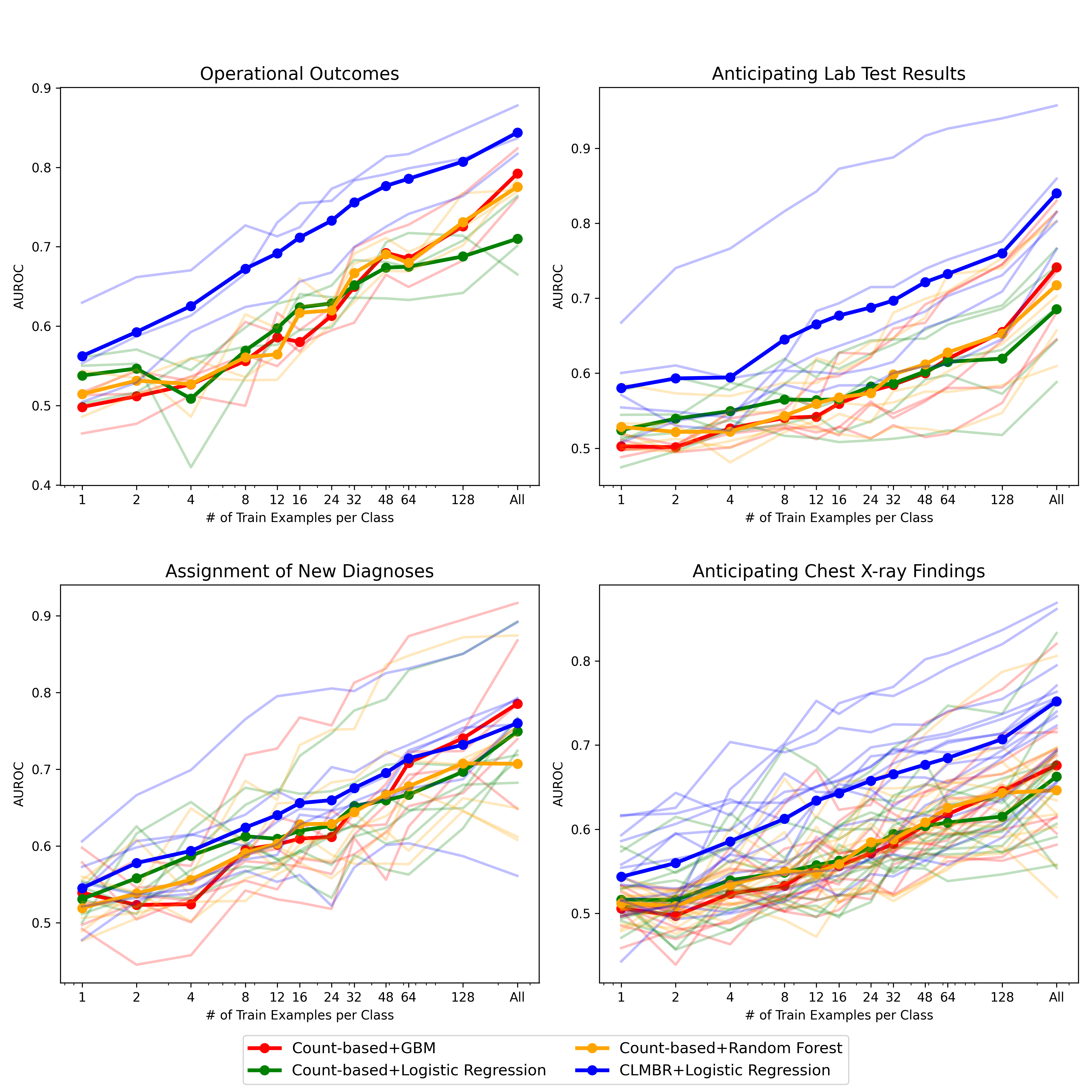}
    \caption{Replication of Figure \ref{fig:agg_auroc} for aggregated AUROC, but including the following baseline models: CLMBR-T-base (\textcolor{blue}{blue}), GBM (\textcolor{red}{red}), Random Forest (\textcolor{darkyellow}{yellow}), and Logistic Regression (\textcolor{darkgreen}{green}).}
    \label{fig:agg_auroc_all}
\end{figure}

\begin{figure}[htbp]
    \centering
    \includegraphics[width=\textwidth]{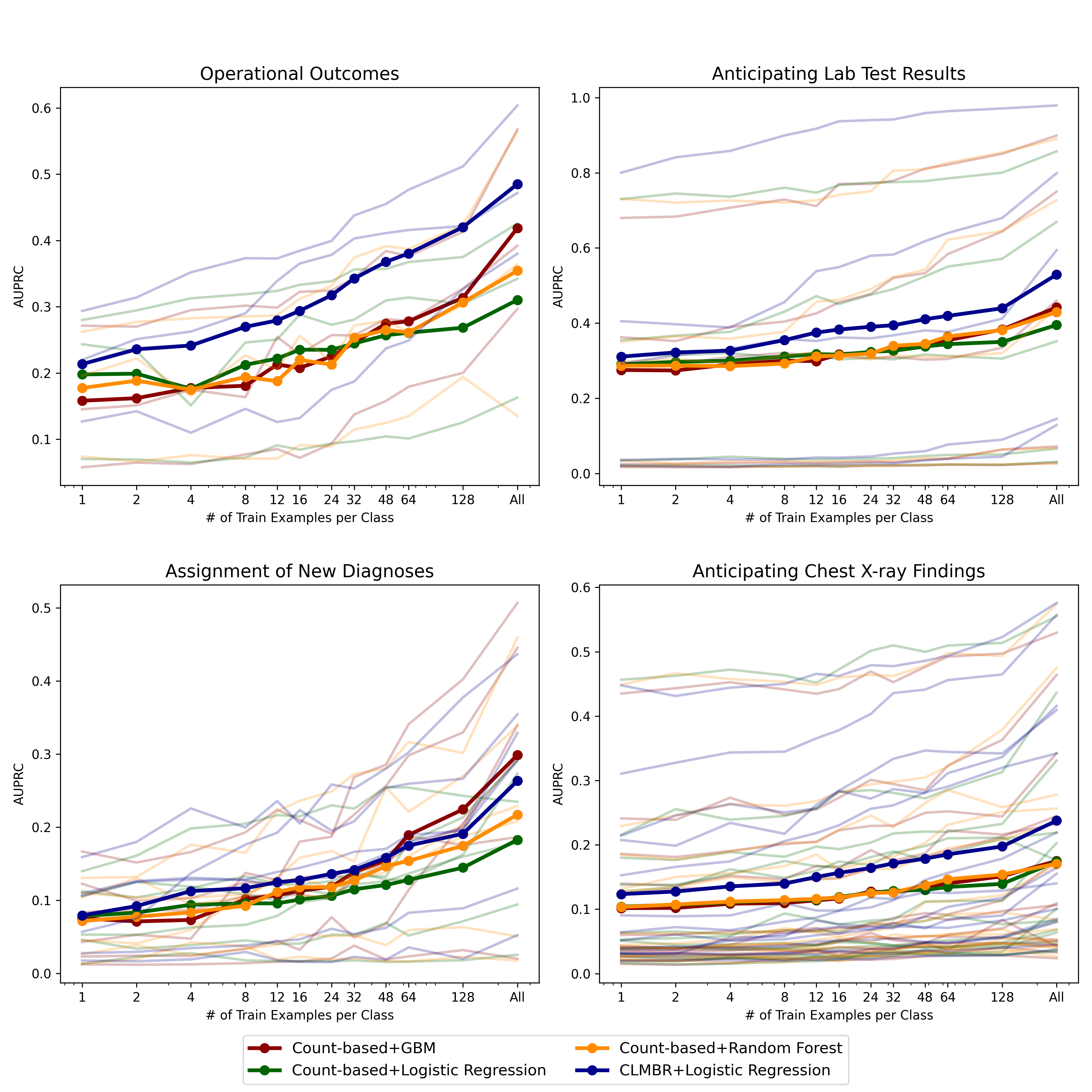}
    \caption{Replication of Figure \ref{fig:agg_auprc} for aggregated AUPRC, but including the following baseline models: CLMBR-T-base (\textcolor{blue}{blue}), GBM (\textcolor{red}{red}), Random Forest (\textcolor{darkyellow}{yellow}), and Logistic Regression (\textcolor{darkgreen}{green})}
    \label{fig:agg_auprc_all}
\end{figure}

\begin{table}[htbp]
  \centering 
  \caption{CLMBR-T-base Hyperparameters}
  \vspace{0.2cm}
  \begin{tabular}{llll}
  \toprule
    \textbf{Name} & \textbf{Values} & \textbf{Best Value}\\
    \midrule
    Learning Rate & 0.0001, 0.00001 & 0.00001 \\ 
    Context Window Size & 496 & 496 \\
    Internal Dropout & 0, 0.2, 0.4 & 0 \\ 
    \# of Layers & 6, 12 & 12 \\ 
    LR Head Learning Rate & 1e-6, 1e-5, ..., 1e5, 1e6 & Task dependent \\ 
    Hidden dimension & 768 & 768\\
    \bottomrule
  \end{tabular}
  \label{tab:clmbr_hyperparameters} 
\end{table}

\begin{table}[htbp]
  \centering 
  \caption{GBM Hyperparameters}
  \vspace{0.2cm}
  \begin{tabular}{llll}
  \toprule
    \textbf{Name} & \textbf{Values} & \textbf{Best Value}\\
    \midrule
    Learning Rate & 0.02, 0.1, 0.5 & Task-dependent \\ 
    Max Depth & 3, 6, -1 & Task-dependent \\ 
    Number of Leaves & 10, 25, 100 & Task-dependent \\ 
    \bottomrule
  \end{tabular}
  \label{tab:gbm_hyperparameters} 
\end{table}

\end{document}